\pdfoutput=1

\documentclass[11pt]{article}

\usepackage[preprint]{acl}

\usepackage{times}
\usepackage{latexsym}
\usepackage{color}  

\usepackage[T1]{fontenc}

\usepackage[utf8]{inputenc}

\usepackage{microtype}

\usepackage{inconsolata}

\usepackage{graphicx}
\usepackage{multirow}
\usepackage{longtable}
\usepackage{booktabs}
\usepackage{xspace}
\usepackage{xcolor}         
\usepackage{subcaption}

\usepackage{epsfig}
\usepackage{graphicx}
\usepackage{float}
\usepackage{wrapfig}
\usepackage{algorithm,algorithmicx,algpseudocode}
\usepackage{bm,xspace}
\usepackage{comment}
\usepackage{verbatim}
\usepackage{multirow}
\usepackage{array}
\usepackage{balance}
\usepackage{url}
\usepackage{booktabs}
\usepackage{etoolbox,siunitx}
\usepackage{calc}
\usepackage{pifont,hologo}
 
\usepackage{nicefrac}

\usepackage{arydshln}
\usepackage[utf8]{inputenc}
\usepackage[T1]{fontenc}
\usepackage{url} 
\usepackage{amsfonts}
\usepackage{nicefrac}
\usepackage{microtype}
\usepackage[american]{babel}
\usepackage{enumitem}
\usepackage{amssymb}
\usepackage{amsmath}
\usepackage{siunitx}
\usepackage{bbm}

\usepackage[super]{nth}
\usepackage{nicefrac}
\robustify\bfseries

\usepackage{dsfont}

\newcommand{\sA}{\mathcal{A}}

\newcommand{\sD}{\mathcal{D}}

\newcommand{\sO}{\mathcal{O}}

\newcommand{\sS}{\mathcal{S}}

\DeclareMathSymbol{@}{\mathord}{letters}{"3B}

\makeatletter
\DeclareRobustCommand\onedot{\futurelet\@let@token\@onedot}
\def\@onedot{\ifx\@let@token.\else.\null\fi\xspace}

\def\eg{\emph{e.g}\onedot} 
\def\ie{\emph{i.e}\onedot} 
 
 \def\vs{\emph{vs}\onedot}

\makeatother

\usepackage{pifont}
\newcommand{\cmark}{\ding{51}}%
\newcommand{\xmark}{\ding{55}}%
\newcommand{\barman}{\textsc{Barman}\xspace}
\newcommand{\blocksworld}{\textsc{BlocksW}\xspace}
\newcommand{\floortile}{\textsc{Floortile}\xspace}
\newcommand{\grippers}{\textsc{Grippers}\xspace}
\newcommand{\storage}{\textsc{Storage}\xspace}
\newcommand{\termes}{\textsc{Termes}\xspace}
\newcommand{\tyreworld}{\textsc{Tyreworld}\xspace}

\newcommand{\planner}{\textsc{Fast-Downward}\xspace}
\newcommand{\validator}{\textsc{VAL}\xspace}

\newcommand{\chatgpt}{\textrm{GPT-3.5}\xspace}
\newcommand{\mistral}{\textrm{Mistral-7B}\xspace}

\newcommand{\ourmethod}{\textrm{Predicting Semantics of Actions with Language Models}\xspace}
\newcommand{\ourmethodshort}{\textrm{PSALM}\xspace}

\newcommand{\actsem}{action semantics generator\xspace}
\newcommand{\actsemshort}{ASG\xspace}

\newcommand{\trajsam}{trajectory sampler\xspace}
\newcommand{\trajsamshort}{TS\xspace}

\title{Language Models Can Infer Action Semantics\\for Symbolic Planners from Environment Feedback}

\author{%
  Wang Zhu \quad\quad Ishika Singh \quad\quad Robin Jia \quad\quad Jesse Thomason \\
  Department of Computer Science\\
  University of Southern California\\
  \texttt{\{wangzhu@usc.edu, ishikasi@usc.edu, robinjia@usc.edu, jessetho@usc.edu\}} \\
}

\begin{document}
\maketitle
\begin{abstract}
Symbolic planners can discover a sequence of actions from initial to goal states given expert-defined, domain-specific logical action semantics.
Large Language Models (LLMs) can directly generate such sequences, but limitations in reasoning and state-tracking often result in plans that are insufficient or unexecutable.
We propose \ourmethod (\ourmethodshort), which automatically learns action semantics by leveraging the strengths of both symbolic planners and LLMs.
\ourmethodshort repeatedly proposes and executes plans, using the LLM to partially generate plans and to infer domain-specific action semantics based on execution outcomes.
\ourmethodshort maintains a belief over possible action semantics that is iteratively updated until a goal state is reached.
Experiments on~7 environments show that when learning just from one goal, \ourmethodshort boosts plan success rate from 36.4\% (on Claude-3.5) to 100\%, and explores the environment more efficiently than prior work to infer ground truth domain action semantics.
\end{abstract}

\section{Introduction}

\begin{figure*}[t]
    \centering
    \includegraphics[width=\textwidth]{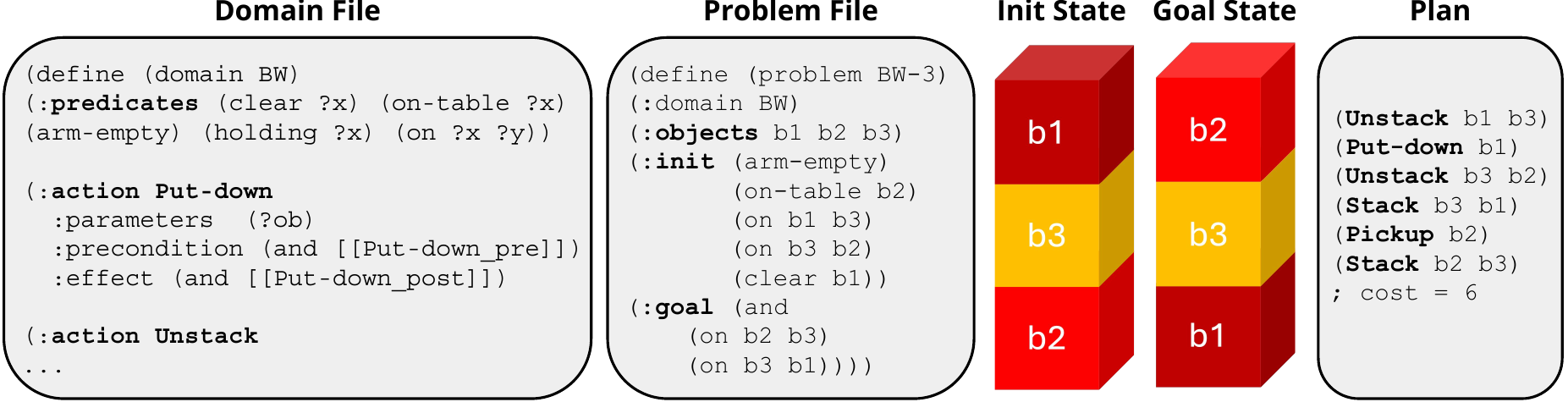}
    \caption{An example of symbolic planning information from the \blocksworld domain,
    from left to right: PDDL domain file, PDDL problem file, visualization of initial and goal state for block stacking, and a potential plan.}
    \label{fig:pddl_dfpf_example}
\end{figure*}

Symbolic planning requires extensive domain knowledge to produce a sequence of actions that achieve a specified goal.
Domain knowledge comprises expert-annotated action semantics that govern the dynamics of the environment.
For example, traditional symbolic methods, like Planning Domain Description Language (PDDL;~\citealp{aeronautiques1998pddl}) solvers, take action semantics annotated in a domain file as input (Figure~\ref{fig:pddl_dfpf_example}).
These symbolic solvers systematically search the state space based on actions that can be executed as per these semantics and return a sequence expected to achieve specified goal conditions, if possible.
However, a human expert must exhaustively define the action semantics of domain to enable symbolic planning.

\begin{figure}
    \centering
    \includegraphics[width=\linewidth]{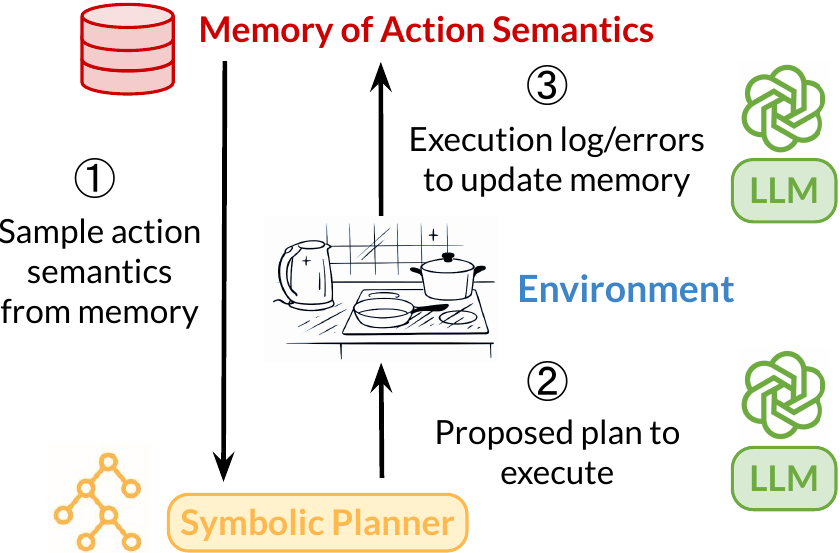}
    \caption{LLMs can propose plans and generate action semantics, but struggle with state tracking.
    Symbolic planners leverage specialized search algorithms, but require predefined action semantics for the environment. 
    \ourmethodshort integrates the strengths of both.}
    \label{fig:teaser}
\end{figure}

We propose a novel domain induction task in which an agent must infer the action semantics of an environment without manual annotation or error correction. 
In this setting, domain information like object properties and action functions headers are given to an agent in natural language.
The agent is then asked to infer the action semantics, that is, symbolic pre- and post-conditions, through interacting with the environment and learning from resulting feedback (Figure~\ref{fig:teaser}).
Our proposed setting is motivated by the longer-term goal of building real-world robots that can explore a new environment (\eg, an apartment) and learn to perform new tasks in the environment (\eg, tidying) by building a symbolic, interpretable world representation.

We draw inspiration from prior work that prompts Large Language Models (LLMs) to perform robotics planning tasks.
These methods rely on LLMs' ability to infer world knowledge from natural language input for generating unseen plans, rather than leveraging expert domain knowledge.
For instance, ProgPrompt~\cite{progprompt} generates program-like high-level plans from LLMs. 

LLMs struggle with long-horizon planning and often generate plans that include invalid actions, such as filling a cup before the lid is removed.
Instead of using the LLM directly as a planner, we propose \ourmethod (\ourmethodshort), a novel method that combines language models with symbolic solvers to \textit{iteratively explore the environment and predict the action semantics based on environment feedback}.
\ourmethodshort maintains a probabilistic memory of learned action semantics, which iteratively improves by interacting with the environment.
We use LLMs to sample possible incorrect or incomplete candidate action trajectories conditioned on initial proposed action sequences from a symbolic planner, then infer action semantics based on the result of executing those trajectories.
\ourmethodshort leverages LLMs' strong commonsense reasoning abilities, as well as their ability to generate syntactically valid formal semantics, while using a symbolic solver to search for ways to achieve the final goal state based on our current belief about the action semantics.

We demonstrate the effectiveness and efficiency of \ourmethodshort for domain induction in 7 symbolic reasoning environments. 
\ourmethodshort can achieve 100\% success rate with the \planner solver, while the best LLMs, O1-preview~\cite{o1-preview} and Claude-3.5-Sonnet~\cite{claude-3.5-sonnet} perform less than 40\% on average.
Additionally, \ourmethodshort consistently induces correct domain files, and does so with substantially fewer total execution steps and environment resets than other approaches.
The integration of LLMs and symbolic solvers is a promising avenue for domain induction in robotics and general reasoning agent workflows.

\begin{table*}[ht]
\centering
\resizebox{\linewidth}{!}{
\begin{tabular}{lcccccc}
    & \bf Goal & \multicolumn{3}{c}{\bf Independent of} & \bf Leveraging & \bf Guaranteed \\
    \cmidrule(lr){3-5}
    & & Partial AS & Valid plans & Human eval & Env feedback & Plan success \\
    \toprule
    \citet{llmp} & Problem File & \xmark & \cmark & \cmark & \xmark & \xmark \\
    \citet{hazra2024saycanpay} & Plan & \cmark & \cmark & \cmark & \xmark & \xmark \\
    \citet{Silver2023GeneralizedPI} & Program & \xmark & \cmark & \cmark & \xmark & \xmark \\
    \midrule
    \citet{arora:hal-02010536} & AS & \cmark & \xmark & \cmark & \xmark & \xmark \\ 
    \citet{Wong2023LearningAP} & AS & \xmark & \cmark & \cmark & \xmark & \xmark \\ 
    \citet{guan2023leveraging} & AS & \cmark & \cmark & \xmark & \xmark & \xmark \\
    \citet{oswald2024large} & AS & \cmark & \cmark & \cmark & \xmark & \xmark \\
    \midrule
    \ourmethodshort & AS & \cmark & \cmark & \cmark & \cmark & \cmark \\
    \bottomrule        
\end{tabular}
}
\caption{Comparing the \ourmethodshort domain induction task setup to representative related works in LLM and LLM-Modulo planning; here, ``AS'' is short for action semantics.}
\label{tab:rw_table}
\end{table*}

\section{Background and Related Works}
\label{sec:rw}
We first introduce symbolic planning, and then compare \ourmethodshort's domain induction setup with previous LLM or LLM-Modulo planning methods, as listed in Table~\ref{tab:rw_table}.

\subsection{Classical and Symbolic Planning}
\label{subsec:classical}
Classical planning algorithms have been widely applied in autonomous spacecrafts, military logistics, manufacturing, games, and robotics. 
The automated STRIPS planner was the first algorithm that operated the Shakey robot~\cite{strips}.
Classical planners require finite, deterministic, and full state information to generate guaranteed plans when a path from the initial to the goal state is possible. 
Some other frameworks were also shown to be useful for robot planning~\cite{prodigy, htn}. 
Symbolic planning languages, such as
Planning Domain Description Language (PDDL; ~\citealp{aeronautiques1998pddl}) and Answer Set Programming (ASP;~\citealp{asp, asp1}), provide a more structured and flexible way to represent problems.

We use PDDL for symbolic planning (Figure~\ref{fig:pddl_dfpf_example}).
We define a planning problem $\mathrm{P}$ as a tuple $\langle\sD, \sO, s^i, \sS^g\rangle$, for $\sD$ the domain, $\sO$ a set of objects in the domain, $s^i$ the initial state, and $\sS^g$ the goal specification, \ie, a set of goal states satisfying the goal conditions. 
A plan solution to $\mathrm{P}$ is a $T$-step sequence of actions $a_{1..T}$, which once executed at $s^i$ would lead to a state in $\sS^g$.

The PDDL domain file also defines the symbolic action set $\sA$ in the environment and \textit{predicatess} representing objects properties.
Each action has a string name (\eg, \texttt{Put-down}), parameters (\eg, \texttt{?ob}), and \textit{semantics}.
The action semantics $\Phi_a$ of an action $a\in\sA$ include the \textit{preconditions} when $a$ is valid to execute, and the \textit{postconditions} (effects) describing what changes when an $a$ is executed. 
Unlike LLM+P~\cite{llmp}, which generates PDDL problem files from in-context examples, we infer action semantics in domain files.

\subsection{LLMs for Task Planning}
Several works utilize LLMs to generate plans directly given initial and goal states~\cite{zeroshot, saycan}, but such stochastic, generative approaches lose the success guarantees of symbolic planners. 
To improve the correctness of the plans, learned latent spaces~\cite{trivedi2021leaps} or LLMs~\cite{progprompt, cap, hu2024deploy} are used to generate executable programs, which introduce some symbolic structure and constraints, reaching up to 90\% plan success on some simple domains, such as \grippers, but these still do not guarantee plan success~\cite{Silver2023GeneralizedPI}.

Recently,~\citet{kambhampati2024position} proposed the LLM-Modulo network, combining LLM plan generation with symbolic planner verifiers.
Using an LLM for final plan generation cannot achieve high success rate in complex domains, such as \barman and \termes~\cite{NEURIPS2023_65a39213, hazra2024saycanpay}.
\citet{guan2023leveraging} have compared LLMs and symbolic planners, finding the latter is consistently better over 3 domains.

In this paper, we consider planning domain induction.
We use LLMs to propose partial plans and predict domain knowledge based on execution feedback.
Once we uncover the domain action semantics, plan success from a symbolic solver is guaranteed, provided the goal is reachable.

\subsection{Domain Induction}
The domain induction problem has been studied under different setups (Appendix Table~\ref{tab:rw_table_appendix}).
\citet{arora:hal-02010536} introduced several setups, including predicting pre- and post-conditions from valid plans or reusable plan fragments.
\citet{Wong2023LearningAP} and~\citet{lasp} filled in missing pre- and post-conditions in the domain file given other action semantics in the same domain.
\citet{guan2023leveraging} generated domain predicates and action semantics using LLMs given detailed text action descriptions. 
Because the generated predicates were not aligned with the problem file, human feedback was required for error correction and evaluation.

Closest to our paper,~\citet{oswald2024large} extended~\citet{guan2023leveraging}'s work by providing predicates and enabling automatic evaluation.
They performed a single-round action semantics generation from LLMs and evaluated the accuracy of the generation.
Using environment feedback, \ourmethodshort effectively solves their domain induction problem, achieving 100\% accuracy over 7 domains.
We advocate for an additional evaluation relevant for downstream, real-world deployment: the efficiency of the environment exploration.
\begin{figure*}[t]
    \centering
    \includegraphics[width=\textwidth]{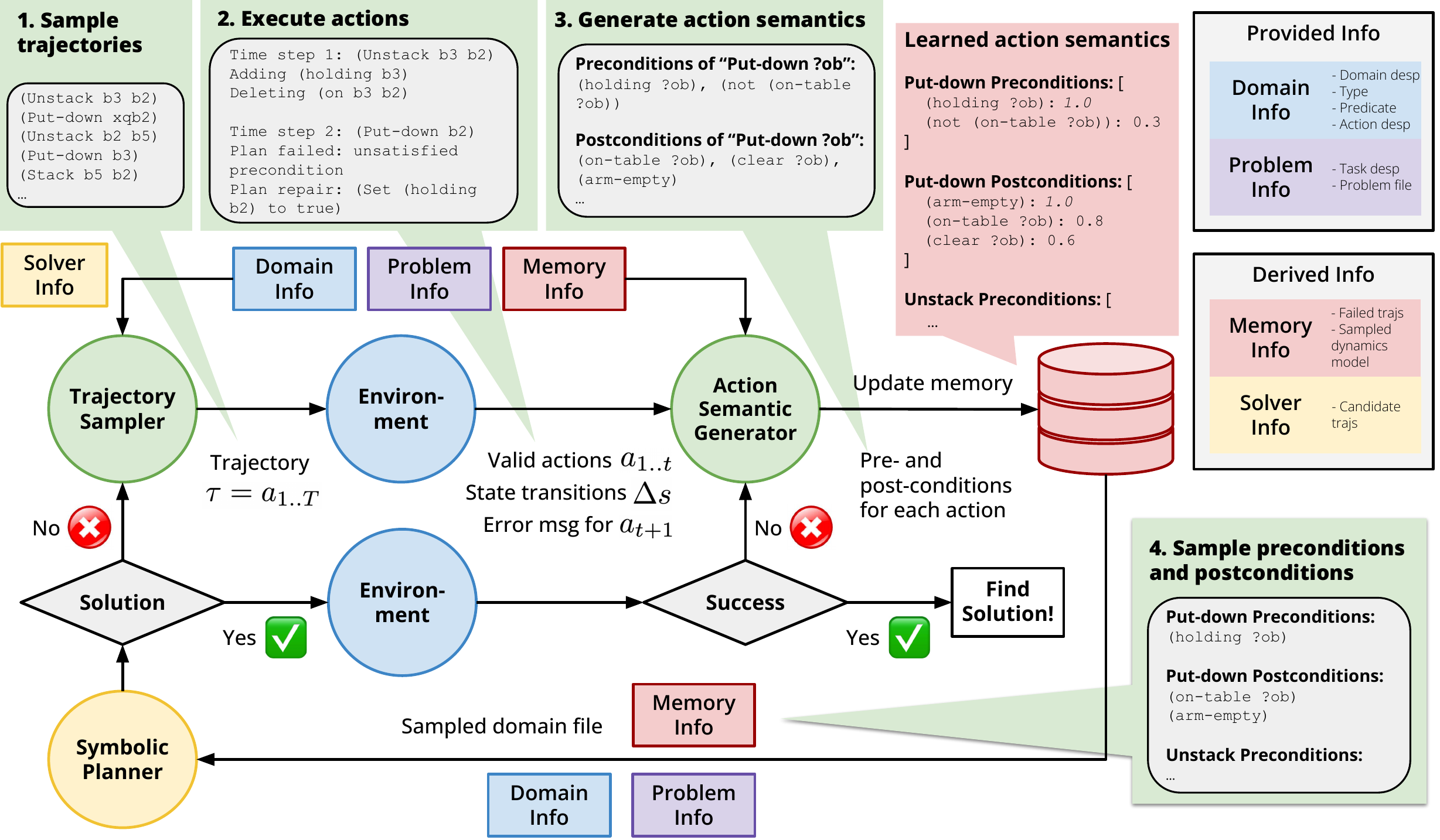}
    \caption{The pipeline of \ourmethodshort in four steps: (1) sample trajectories from a trajectory sampler; (2) execute the trajectories in the environment to get feedbacks (3) generate action semantics for each action with environment feedback, and update the memory based on the prediction; (4) sample action semantics from the memory to construct the domain file for the symbolic solver to check the success.}
    \label{fig:method}
\end{figure*}

\section{Problem Formulation}
Symbolic planning in PDDL requires a domain file characterizing the environment.
Human experts must carefully annotate the action semantics to enable the symbolic solver to find correct solutions.
We propose a novel \textit{domain induction} task, where AI agents must find the action semantics for a new domain without human annotation or correction.

In the domain induction task, the agent knows $\sD \setminus \bigcup_{a\in\sA}{\Phi_a}$, including the predicates and the action names, but not the action semantics.
In addition, the agent has access to one problem file $\langle\sO, s^i, \sS^g\rangle$.
The goal of the agent is to learn the correct action semantics $\Phi_a$ for each action $a$.
During the learning process, the agent interacts with the environment initialized to state $s^i$ with an open-loop execution of planned actions, after which the environment is reset back to $s^i$. 
After learning action semantics, the agent should be able to leverage a symbolic solver for efficient and robust task solving in the domain to achieve a state in $\sS^g$.

We evaluate the domain induction learning process on three measures.
(1) Accuracy (Acc): $\sum_{a\in\sA}|\hat{\Phi}_a\cap\Phi_a|/\sum_{a\in\sA}|\Phi_a|$, where $\hat{\Phi}_a$ is the predicted action semantics for action $a$. 
Finding a path to $\sS^g$ is not sufficient to imply correct action semantics, but recovering ground truth action semantics is sufficient to reach $\sS^g$ via the symbolic solver.
(2) The number of resets (NR): how many times the environment returns to $s_i$ because planned actions did not reach $\sS^g$.
(3) The number of executed steps (NES): how many total actions were executed during learning, including those causing failure.

\begin{table*}[ht]
\centering
\tabcolsep 2pt
\begin{tabular}{lrrrrrrr}
    Model & \barman & \blocksworld & \floortile & \grippers & \storage & \termes & \tyreworld \\
    \toprule
    \multicolumn{7}{l}{\textit{Direct plan generation}}\\
    O1-preview & 0 & 30 & 0 & 90 & 5 & 0 & 30 \\
    GPT-4-Turbo & 0 & 20 & 0 & 55 & 5 & 0 & 70 \\
    Claude 3.5 Sonnet & 0 & 60 & 0 & 90 & 5 & 0 & 100 \\
    \midrule
    \multicolumn{7}{l}{\textit{Domain induction + Symbolic planner}}\\
    GPT-4-Turbo & 0 & 0 & 0 & 0 & 5 & 0 & 0 \\
    \citet{oswald2024large} & 0 & 0 & 0 & 0 & 5 & 0 & 0 \\
    PSALM & \bf 100 & \bf 100 & \bf 100 & \bf 100 & \bf 100 & \bf 100 & \bf 100 \\
    \bottomrule        
\end{tabular}
\caption{Plan success rate over 20 examples, comparing \ourmethodshort with LLM and domain induction baselines.}
\label{tab:baselines}
\end{table*}

\section{Proposed Approach: \ourmethodshort}

We propose the \ourmethod (\ourmethodshort) framework, which leverages LLM commonsense, formal planning, and environment feedback.

\subsection{Overview}

Given a goal,
we first use an LLM as a \textit{\trajsam}, and execute the trajectories in the environment to get feedback (Figure~\ref{fig:method}). 
We use the LLM again, together with a rule-based parser, as the \textit{\actsem} to predict action semantics, \ie, preconditions and postconditions, for each action based on that feedback. 
We update the memory of the learned action semantics, ten sample that memory for hypothesized action semantics used as input to the symbolic solver.
Finally, the symbolic solver tries to generate a plan from the goal and hypothesized action semantics.

If the symbolic solver finds a solution, we execute the plan in the environment. 
If the plan reaches the goal, we finish the loop; otherwise, we will pass the result of the failed plan to the LLM to predict the action semantics again.
If the symbolic solver does not find a solution, we will provide some partial candidate trajectories from the symbolic solver to the \trajsam as a prefix sequence of actions for the plan to be generated.


\subsection{Trajectory sampler}
We prompt an LLM with domain and problem information (as described in \S\ref{subsec:classical}), memory information representing our current beliefs about action pre- and post-conditions, and partial trajectories generated by the solver. 
We use templates to convert these into a natural language prompt.
For example, the postconditions \texttt{(on-table ?ob) (arm-empty)} of the action \texttt{put-down} are converted to natural language \textit{The effects are (on-table ?ob), (arm-empty).}
Because the symbolic solver we use, \ie, \planner, is performing a search algorithm given a time-limit of $W$,
if the solver finds any candidate trajectories, \textit{i.e.}, not a complete solution but a partial solution stopped by the timer or a dead end, we include the $k$ longest candidate trajectories as additional input to the \trajsam.
We filter out invalid candidate trajectories that failed in previous iterations.
The \trajsam prompt finally specifies to pick one of the candidate trajectories and generate a trajectory starting from that. 
We execute $l$ trajectories sampled in this way from the.
When $l>1$, we predict action semantics for each trajectory separately (\S\ref{subsec:asg}).
After execution, we update the semantic hypothesis memory (\S\ref{subsec:memory}).

\paragraph{Prospection.} 
Although the LLM's prompt includes the current beliefs about action preconditions, the LLM can still generate trajectories that violate these preconditions. 
We add \textit{trajectory prospection}, enabling the system to do forward prediction in an open-loop fashion before actually invoking the simulator based on its symbolic understanding of the world.
For a generated trajectory $\mathbf{\tau}=a_{1:T}$, we check if is aligned with the current belief of the environment action semantics for $v$ steps.
For $v\le T$, if any action in $a_{1:v}$ does not satisfy the preconditions in the sampled action semantics, starting from that action, we will keep randomly sampling one action until the sampled action is valid in the hypothesized action semantics, repeated to $v$ total actions. 
Otherwise, if $a_{1:v}$ are all valid, we execute the original $a_{1:T}$.

\paragraph{Random sampler.}
We create an ablated \ourmethodshort that randomly samples $v$ actions per trajectory from the longest candidate trajectory, if available, from the symbolic solver during each iteration. 
This random sampler ablation can also take advantage of the prospection module.

\begin{table*}[ht]
\centering
\tabcolsep 2pt
\begin{tabular}{lrrrrrrr}
    Model & \barman & \blocksworld & \floortile & \grippers & \storage & \termes & \tyreworld \\
    \toprule
    GPT-4-Turbo & 53 & 61 & 58 & 70 & 45 & 41 & 67 \\
    \citet{oswald2024large} & 65 & 67 & 58 & 75 & 58 & 59 & 80 \\
    PSALM & \bf 100 & \bf 100 & \bf 100 & \bf 100 & \bf 100 & \bf 100 & \bf 100 \\
    \bottomrule        
\end{tabular}
\caption{Action semantics accuracy over 20 examples, comparing \ourmethodshort with LLM to baselines.}
\label{tab:baselines_as}
\end{table*}

\subsection{Action semantics generator}
\label{subsec:asg}
We combine LLM-based and rule-based action semantics generation given environment feedback.

\paragraph{LLM-based generator.}
We prompt the LLM to generate the action semantics of each action separately.
The prompt contains information about the domain, problem, memory information, and environment feedback.
For a trajectory $\mathbf{\tau}=a_{1:T}$ failed at step $t+1$, environment feedback for the LLM \actsem takes the form of valid actions $a_{1:t}$, state transitions $s_{j+1}-s_{j}, j=1, \dotsc, t$, where $s_1$ is the initial state $s^i$, and the environment error message string for action $a_{t+1}$.
Note that each state is a set of predicate assignments, so we can compute the difference of state transition as the difference of two sets.
We assume the error message is provided by the environment. 
In the tested environments, when an action fails the error message specifies an unsatisfied precondition.
When no action fails, the error message either states that the the goal is not reached, or the goal is reached and the \ourmethodshort\ inference loop exits.
For each action, we prompt the LLM once to predict the preconditions and once for the postconditions.

\paragraph{Rule-based generator.}
We also write a rule-based parser for the output from the environment feedback to infer one or more missing preconditions suggested by error messages.
Rule-based parser also scan state transition descriptions to derive postconditions.
For each iteration, we update the action semantics hypothesis memory with both LLM-generated pre- and post-conditions and those from this rule-based generator.
Unlike the LLM \actsem, the rule-based generator relies solely on the feedback at the current time step, making it less efficient.

\subsection{Memory of action semantics}
\label{subsec:memory}
We keep a memory of the hypothesized action semantics. 
For each action's action semantics, we store two lists of predicted statements for preconditions and postconditions.
Each statement $\phi$ is associated with a belief $p(\phi\in\Phi_a|a)$, in short $p(\phi|a)$. 
We use this belief as a binary probability for each statement to sample, at each iteration, the concrete action semantics input to the symbolic solver.

The first time a statement $\phi$ is predicted as part of the action semantics for $a$, the belief will be assigned to $1$. 
Afterwards, this belief will be updated following an exponential forgetting rule.
Suppose at time step $t$, the predicted action semantics of $a$ is $\hat{\Phi}_{a, t}$, the belief update rule of statement $\phi$ is
\begin{align*}
    &p_{t+1}(\phi|a) = \\
    &\begin{cases}
    \mathbbm{1}[\phi\in \hat{\Phi}_{a, t+1}] \qquad \qquad\qquad \text{if } p_{t}(\phi|a)=0 \\
    \gamma_\phi p_{t}(\phi|a) + (1-\gamma_\phi) \mathbbm{1}[\phi\in \hat{\Phi}_{a, t+1}] \quad \text{else} \\
    \end{cases}
\end{align*}
where $\gamma_\phi$ is the forgetting factor.
If a statement $\phi$ has only been predicted by the LLM. 
Once a statement $\phi$ is predicted by rule-based generator, $\gamma_{\phi}=1$, which does not decay.
Notice that all the statements predicted in the current time step and in the memory $\phi\in\bigcup_{t'\in\{1..t+1\}}\hat{\Phi}_{a, t'}$ will be updated following the rule.

\begin{figure*}
    \centering
    \includegraphics[width=\textwidth]{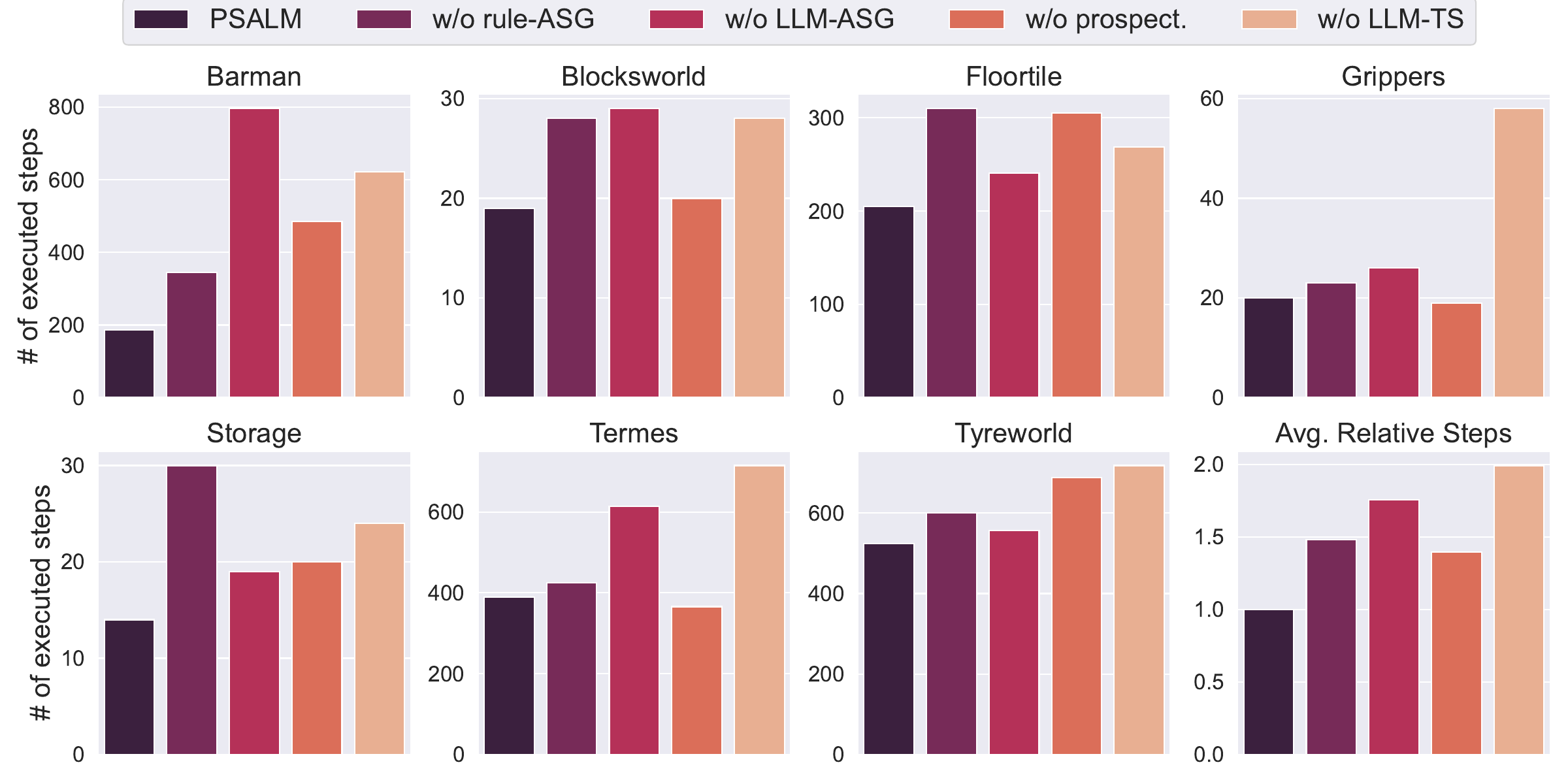}
    \caption{We compare \ourmethodshort with multiple variations over 7 domains. We report on NES and the results suggest (1) LLM as a \trajsam greatly reduces the execution steps; (2) LLM and rule-based action semantics generators have complementary benefits; and (3) Prospection to reject trajectories based on current action semantics hypotheses is helpful overall. \trajsamshort is short for \trajsam and \actsemshort is short for \actsem.}
    \label{fig:main_results}
\end{figure*}

\section{Experiments}

We show that \ourmethodshort can recover 100\% the action semantics in 7 symbolic domains, resulting in 100\% planning success rate as well. 
The critical advantage the LLM component of \ourmethodshort\ is to greatly reduce the cost, \ie, the number of resets and the number of executed steps, for achieving the iterative recovery of these domain semantics.

\subsection{Experimental setups}
We experiment on 7 symbolic domains from International Planning Competitions~\cite{seipp-et-al-zenodo2022}; each defines 20 tasks that vary in number of environment objects and optimal plan length.
{\bf (1) \barman}: The robot is a bartender with 2 hands preparing cocktails for a customer’s order, using the specified ingredients and appropriate tools; 
{\bf (2) \blocksworld}: The robot reorganizes a collection of block piles arranged on a table, into a specified configuration while adhering to the simple physics principles;
{\bf (3) \floortile}: A set of robots painting
color patterns on floor tiles, allowed to move around but not to step on painted tiles;
{\bf (4) \grippers}: A set of robots with 2 grippers each are given a task to move objects among different rooms;
{\bf (5) \storage}: The robot lifts and drops crates initially stored in different areas, into a depot, using a given set of hoists;
{\bf (6) \termes}: The robot constructs complex structures by transporting and positioning blocks, as well as using them as a means to move adjacent blocks;
{\bf (7) \tyreworld}: The robot is assigned with changing flat tires, which involves tasks such as removing flat tires, inflating the intact tires, tightening nuts, and returning tools to the boot.

In each domain, we choose the first task that requires using all actions to reach the goal, learn action semantics from that task, then test on all 20 tasks. 
All experiments use the \planner~\cite{fast-downward} planner, with a search time limit of $W=30$ seconds during the \ourmethodshort loop, and unlimited search time during testing (Table~\ref{tab:baselines}).
We use \validator\footnote{https://github.com/KCL-Planning/VAL} as simulation environment for plan validation, calculating state condition changes, and obtaining error messages.
We set the maximum number of \ourmethodshort induction iterations to 1k for pure random baselines and to 100 for any method involving LLMs.
We report the average over 3 runs for all the methods.

For the main results, we use GPT-4-Turbo~\cite{openai2023gpt4} as the language model agent, with temperature 0 and one-shot prompting following~\citet{llmp}.
We detail prompts in Appendix~\ref{sec:app_prompt}.
We use $v=10$ prospection steps, $l=1$ sampled trajectory,  $k=3$ candidate trajectories, $g=5$ failed trajectories per iteration, $\gamma_\phi=0.8$ memory forgetting factor.
We perform ablation studies over these hyperparameters in the analysis.
More experimental details are in Appendix~\ref{appsec:exp_detail}.

\subsection{Comparison with baselines}
As shown in Table~\ref{tab:baselines}, \ourmethodshort achieve 100\% planning success rate, outperforming multiple LLM direct plan generation baselines. Other domain induction baselines using LLMs, such as~\citet{guan2023leveraging}, are unable induce correct action semantics because they do not leverage environment feedback.
Direct prompting for the plan of the task, even with powerful LLM models, O1-preview and Clalude-3.5 Sonnet, cannot solve any task in the complex domains such as \barman or \termes.

Table~\ref{tab:baselines_as} shows that without environment feedback, GPT-4 and \citet{guan2023leveraging} can predict part of the action semantics of the environment, but cannot recover the full action semantics correctly.
Given only partially correct action semantics, the solver will either fail to find a plan, or generate a plan that fails when executed.
We provide the prompt templates for plan generation and domain induction as in Appendix~\ref{sec:app_prompt}.

\subsection{Ablations on using LLM and prospection}

Figure~\ref{fig:main_results} compares \ourmethodshort with multiple baselines over 7 domains, including no rule-based \actsem (w/o rule-ASG), no LLM \actsem (w/o LLM-ASG), no prospection (w/o prospect.) and use random trajectory sampler (w/o LLM-TS).
Each achieves 100\% Acc, so we focus on number of executed steps (NES).

\paragraph{LLM as a sampler greatly reduces the execution steps.}
We find that the LLM \trajsam reduces the execution steps over all domains.
The random sampler, even with prospection, lacks the commonsense reasoning knowledge for choosing trajectories likely to be solutions to the problem.

\paragraph{LLM and rule-based action semantics predictions have complementary benefits.}
Comparing the purple bars (w/o rule-\actsemshort) and the magenta bars  (w/o LLM-\actsemshort), rule-based \actsem is more important in \floortile, \storage and \tyreworld, while LLM \actsem is more important in the other four domains.
With the commonsense prior knowledge provided by the LLM, and the exactness of the information from the rule-based parser, combining both predictions is always a better solution.

\paragraph{Prospection induces redundant actions sometimes, but is necessary overall.}
For certain domains with a small number of actions $|\sA|$, such as \grippers ($|\sA|=3$) and \termes ($|\sA|=7$), a large number of prospection steps slows search.
However, prospection is overall helpful.
The w/o prospect. bar shows around 1.4 times execution steps on average \vs \ourmethodshort.
Especially, in the \barman domain, prospection reduces more than a half of the execution steps.

\begin{figure*}
    \centering
    \includegraphics[width=\textwidth]{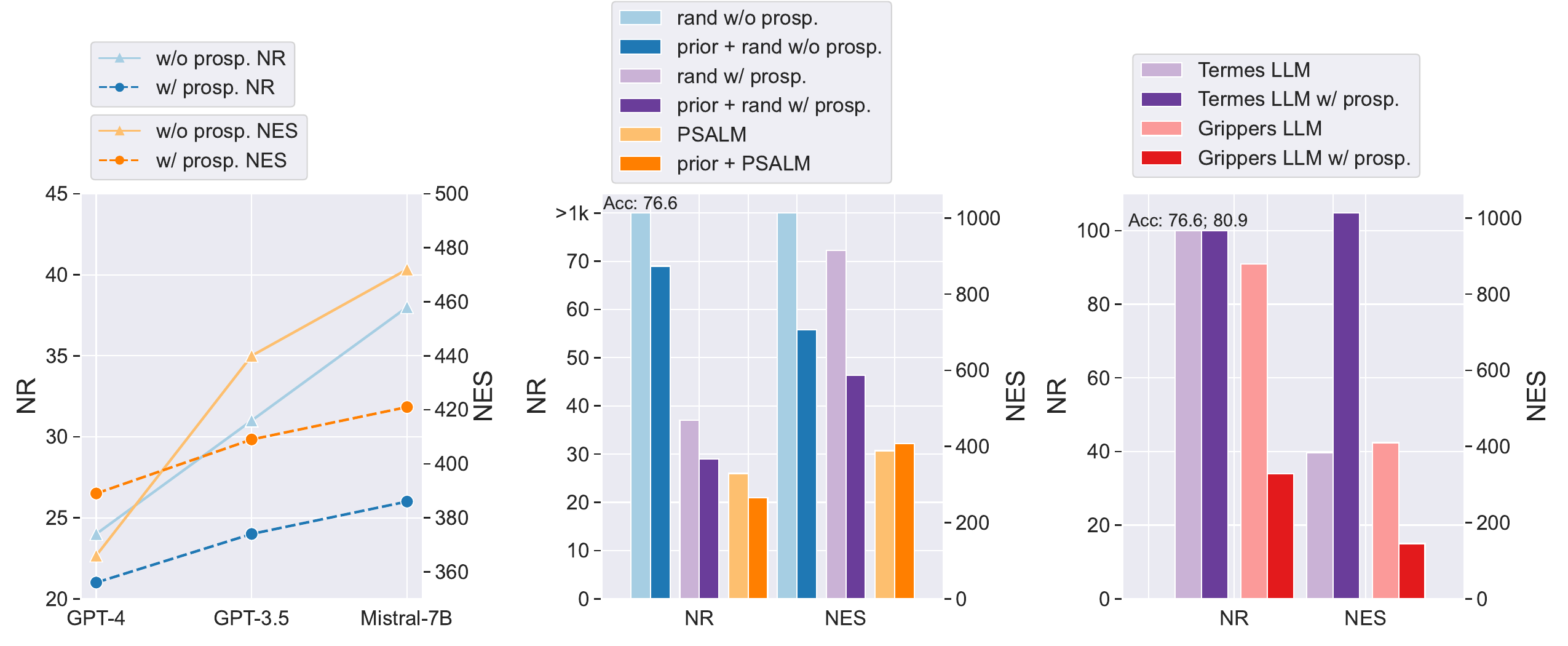}
    \caption{Additional analysis for PSALM. (Left) We vary the type of LLM and show that \ourmethodshort works with \chatgpt and \mistral on the \termes domain. (Middle) Using the LLM prior before trajectory sampling (darker bars) enables the random baselines to work better compared to not having the prior (lighter bars), though it can adversely affect the full \ourmethodshort method. (Right) Experiments where we remove the error message from input to the LLM \actsem. Without error messages, \ourmethodshort works only on easy domains. For the experiments that fail to find a solution of the problem, we show the action semantics accuracy on top of the bar.}
    \label{fig:analysis}
\end{figure*}
\subsection{Analysis}

\paragraph{\ourmethodshort can work on less powerful and public LLMs.}
Figure~\ref{fig:analysis} left shows \ourmethodshort can work with 100\% plan success rate on \chatgpt and \mistral, though with more resets (NR) and number of execution steps (NES).
Prospection reduces the gap between the different LLMs.

\paragraph{LLM prior enables random baselines.}
We experiment with using the LLM to predict the action semantics $\hat{\Phi}_{0, a}$ for each action $a$, before having any sampled trajectory, and then start the iteration with the memory of the dynamics model initialized from $\hat{\Phi}_{0, a}$.
This prediction is solely based on the name of the action and its parameter.
Figure~\ref{fig:analysis} middle shows on the \termes domain the LLM prior can enable the random baseline without prospection to work within 1k resets, and greatly reduce NR and NES for the random baseline with prospection.
However, on more complex domains such as \barman, the LLM prior does not enable the random baselines, \ie, failing to find a solution within 1k resets, with and without prospection.
The finding implies we can perform domain induction on some easier domains with limited access, or even one call, of language models.
On the other hand, this naive LLM prior is harmful to \ourmethodshort, because it inserts noisy predictions of action semantics which requires additional resets to erase.

\paragraph{Error messages are crucial for domain induction.}
Error messages are obtainable from simulation environments, or in real world deployment if human operators are available to describe failure.
We study whether \ourmethodshort can succeed without receiving error messages about failed actions.
Note that without the error message, the \actsem can only be LLM-based and cannot utilize rule-based inference.
Figure~\ref{fig:analysis} right shows that on slightly complex domains like \termes, \ourmethodshort cannot predict the correct dynamics model, \ie, the accuracy is not 100. 
\ourmethodshort works on very simple domains like \grippers, but it requires 7x more NES to learn.
Moreover, prospection is essential when we have no error message.
\section{Conclusion}

In conclusion, we propose a novel domain induction problem in PDDL, where agents automatically infer the action semantics for a new domain without human annotation.
We introduce a simple but strong framework \ourmethodshort, which combines the commonsense reasoning ability of large language models with the precision of symbolic solvers for domain induction. 
To update the action semantics in a memory, \ourmethodshort uses LLMs as agents for trajectory sampling and action semantics prediction.
We demonstrate the effectiveness and efficiency of leveraging LLMs with environment feedback for domain induction over 7 domains.

\section*{Limitations}

Though \ourmethodshort shows to be relatively robust across multiple LLMs, we do not claim that \ourmethodshort can work with \textit{any LLM}.
Additionally, the error message, which might be hard to get from some real-world environment, is crucial in the \ourmethodshort framework given current LLM reasoning abilities.
Future works may explore methods without the error message as the input.
To apply the domain induction setup to the real world, the environment predicates and object types have to be annotated or derived from the environment. The domain induction problem without these annotations is more challenging, but expert definitions of these domain components represent substantially less effort than fully specifying action pre- and post-conditions. 
We will leave this extension for future studies.

\section*{Acknowledgments and Disclosure of Funding}
WZ and RJ were supported in part by a grant from Open Philanthropy.
IS and JT were supported in part by a grant from the Army
Research Lab (ARL) Army AI Innovations Institute (A2I2),
award number W911NF-23-2-0010.

\bibliography{ref}

\clearpage
\appendix

\section{More Related Works}
As shown in Table~\ref{tab:rw_table}, the previous works have not leveraged environment feedback and guaranteed plan success as \ourmethodshort.
To discuss the related works comprehensively in using LLM in planning and PDDL generation over many setups, we extend the related work as in Table~\ref{tab:rw_table_appendix}.

\subsection{LLM as an Idea Generator in Planning}
LLMs are widely used as idea generators, not only in planning, but also in other reasoning areas, such as first-order logic~\cite{olausson-etal-2023-linc}, multistep reasoning~\cite{zhu-etal-2023-chain}, decomposition~\cite{prasad-etal-2024-adapt} and formal language~\cite{mavrogiannis2023cook2ltl}.

The major difference between LLM idea generators in planning is whether LLMs generate ideas on the plan itself or on the action semantics.
Previous works explore different ways of using LLMs to generate the plan itself, such as Monte Carlo tree search over possible plans~\cite{hazra2024saycanpay}, or applying other verifier or postprocessor after the plan generation~\cite{lin2024clmasp}.
These approaches are better than the direct prompting using LLM, while they still cannot guarantee plan success.

Alternatively, using LLMs to generate action semantics followed by a symbolic planner often results in an all-or-nothing issue during plan generation. Either the action semantics are fully captured, leading to a successful plan, or the planner fails to produce a correct plan for the domain at all. 
Most previous works focus on error analysis.
For instance, ~\citet{pddlgendomain} attempts to generate syntactically correct but not semantically guaranteed PDDL domain using LLMs, by exhaustively listing syntax errors in the prompt to correct the generated domains.
~\citet{oswald2024large} utilizes LLM to generate PDDL actions given detailed text action descriptions and analysis of the errors.
Other works like InterPreT~\cite{han2024interpret} leverages language feedback from human during embodied interaction for domain predicates and action semantics prediction.
They explore complex visual environments and show how to accomplish certain human specified goal with robots, such as ``stack red block on coaster''.
Unlike all previous works, \ourmethodshort shows how we can achieve 100\% success rate on a certain type of task in the domain, by using LLMs to learn the action semantics from one example.

\subsection{LLM for PDDL Generation}
There are two types of PDDL generation for LLM, generating the PDDL problem file or the domain file.

Generating the problem file requires to get access to the domain file, recent works such as LLM+P~\cite{llmp}, PDDLEGO~\cite{zhang-etal-2024-pddlego} demonstrate the effectiveness of LLM to generate problem file for simple domains. The Planetarium~\cite{zuo2024planetariumrigorousbenchmarktranslating} benchmark provides large-scale study of the LLM capability in generating PDDL problem files.

Unlike the success on generating problem file, LLMs, even the most powerful ones, performs badly on generating domain file~\cite{pddlgendomain, oswald2024large}.
We formally define the domain induction problem, its evaluation measures, and provide a first-step solution \ourmethodshort to predicting the action semantics in the domain file in the text-based environments.

\subsection{LLM-based Memory Augmentation}
Knowledge acquisition for task planning through dialog and web access has been studied in the past. 
Previous works~\cite{KnoWDiaL, augknowledge} construct a knowledge base in an open domain to refer to for grounding user utterance. 
Recent works have shown LLM based memory augmentation to be effective.
~\citet{memAug} builds a memory of language instruction and corresponding plan to retrieve from for prompting the LLM with a new instruction, where the retrieved interaction might inform planning.
\ourmethodshort build probabilistic memory storing the belief for domain action semantics prediction.

\begin{table*}[ht]
\centering
\resizebox{\linewidth}{!}{
\begin{tabular}{lcccccc}
    & \bf Goal & \multicolumn{3}{c}{\bf Independent of} & Role of LLM & Final planner \\
    \cmidrule(lr){3-5}
    & & Partial AS & Valid plans & Human eval \\
    \toprule
    \citet{llmp} & PF & \xmark & \cmark & \cmark & PF generator & - \\
    \citet{zhang-etal-2024-pddlego} & PF & \cmark & \cmark & \cmark & PF generator & - \\
    \citet{hazra2024saycanpay} & Plan & \cmark & \cmark & \cmark & Plan sampler & Symbolic + LLM \\
    \citet{valmeekam2023on} & Plan & \cmark & \cmark & \xmark & Plan sampler & LLM \\
    \citet{lin2024clmasp} & Plan & \cmark & \cmark & \xmark & Plan sampler & Symbolic + LLM \\
    \citet{Silver2023GeneralizedPI} & Program & \xmark & \cmark & \cmark & Code generator & LLM \\
    \midrule
    \citet{arora:hal-02010536} & AS & \cmark & \xmark & \cmark & - & - \\ 
    \citet{Wong2023LearningAP} & AS & \xmark & \cmark & \cmark & AS generator & Symbolic + LLM \\ 
    \citet{guan2023leveraging} & AS & \cmark & \cmark & \xmark & AS generator & Symbolic / LLM \\
    \citet{oswald2024large} & AS & \cmark & \cmark & \cmark & AS generator & Symbolic \\
    \multirow{2}{*}{\citet{han2024interpret}} & \multirow{2}{*}{AS} & \multirow{2}{*}{\cmark} & \multirow{2}{*}{\cmark} & \multirow{2}{*}{\xmark} & PF generator + & \multirow{2}{*}{Symbolic} \\
    & & & & & AS generator \\
    \midrule
    \multirow{2}{*}{\ourmethodshort} & \multirow{2}{*}{AS} & \multirow{2}{*}{\cmark} & \multirow{2}{*}{\cmark} & \multirow{2}{*}{\cmark} & Plan sampler + & \multirow{2}{*}{Symbolic} \\
    & & & & & AS generator \\
    
    \bottomrule        
\end{tabular}
}
\caption{Comparing the \ourmethodshort domain induction task setup to representative related works in LLM and LLM-Modulo planning; here, ``PF'' is short for problem file and ``AS'' is short for action semantics. We consider the final planner using tree or heuristic search as symbolic.}
\label{tab:rw_table_appendix}
\end{table*}

\section{Experiment Details}
\label{appsec:exp_detail}

\subsection{Task Selection and Domain Continual Learning}
For learning action semantics, we choose the first task that requires using all actions to reach the goal. Specifically, for \floortile, \barman, \termes and \tyreworld domains, we choose task \#1. For \grippers and \blocksworld domains, we choose task \#2. For \storage domain, we choose task \#3, out of the 20 tasks.

If \ourmethodshort learns from a task that only requires partial actions to reach the goal. 
The planner may be able to generate successful plans for that tasks without knowing the entire domain.
To overcome this issue, we also experiment on a setup where \ourmethodshort learns from task \#1 for all domains and perform a continual learning setup.
We report the learning process as below.
Instead of directly learning the entire \storage domain from task \#3 with 14 executed steps, we start the learning from task \#1 with 13 executed steps, and recover 78.9\% of the action semantics conditions. 
Then, based on what we learned from task \#1, it takes 10 steps to recover 81.6\% of the action semantics conditions learning from task \#2, and 11 more steps to recover the entire domain from task \#3. 
The continual learning process can be applied to real world exploration of certain domain as well. 
Thus, the ``success without learning the entire domain'' effect is not a limitation but a benefit of \ourmethodshort.

\subsection{Task Example Initial and Goal States}
We provide the visualization of the initial and goal state for the selected training example per domain. The natural language description of the actions are from~\citet{llmp}.

\begin{figure*}[ht]
    \centering
    \setkeys{Gin}{width=\linewidth}
    \begin{subfigure}[b]{0.5\textwidth}
    \hfil
    \includegraphics{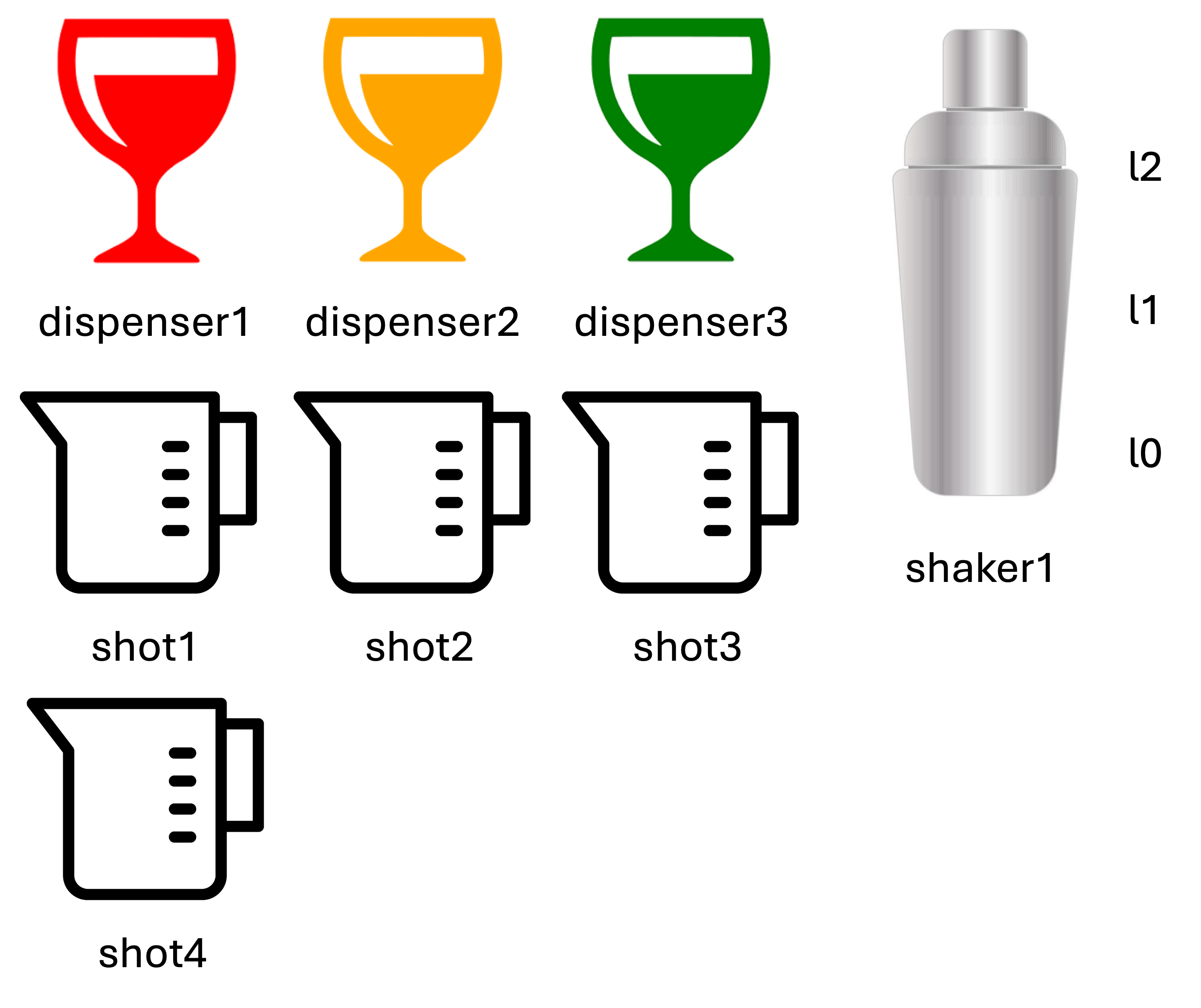}
    \caption{}
    \label{fig:barman_init}
    \end{subfigure}
    \hfil
    \begin{subfigure}[b]{0.35\textwidth}
    \includegraphics{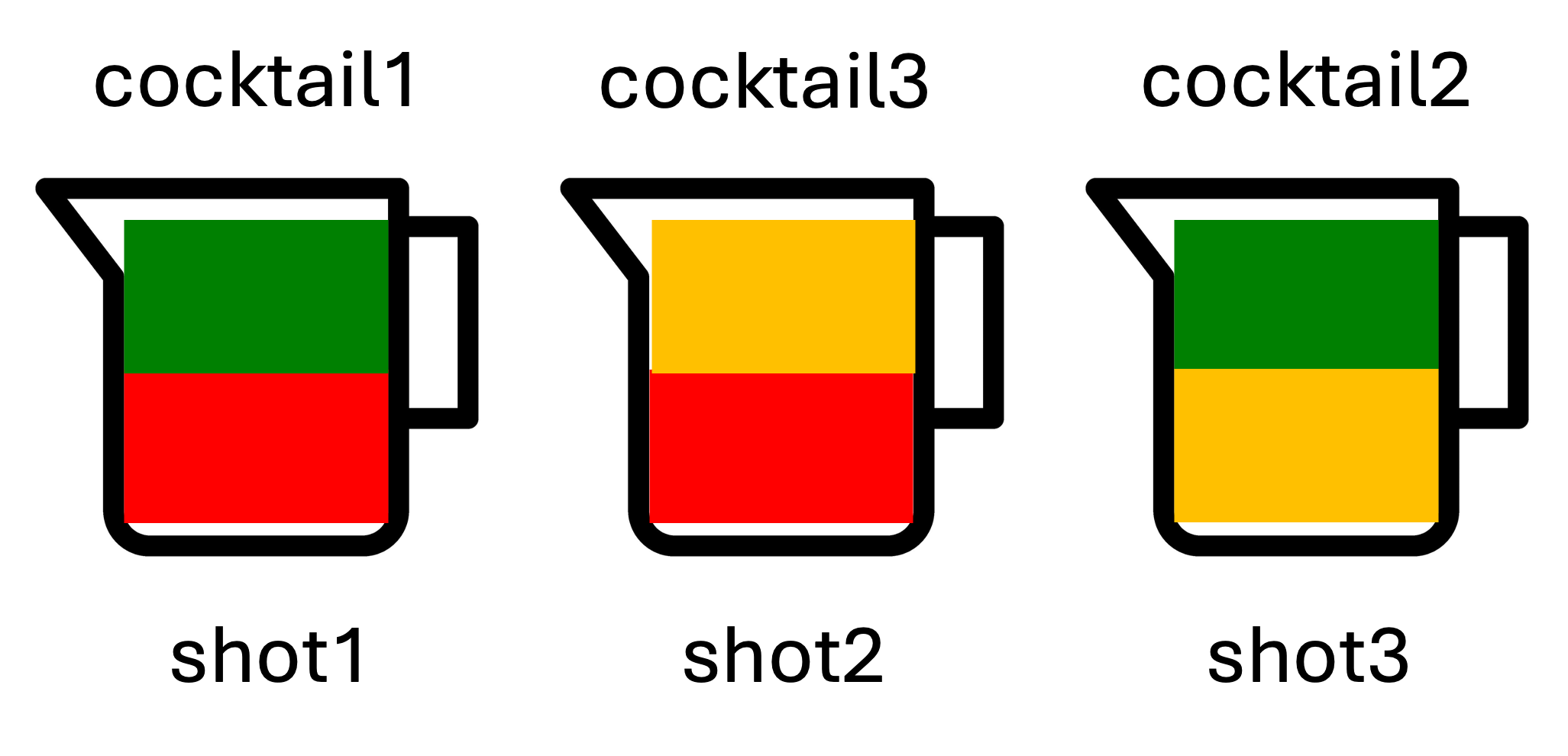}
    \caption{}
    \label{fig:barman_goal}
    \end{subfigure}
    \caption{\barman task \#1 initial state (a) and goal state (b).}
    \label{fig:barman_example}
\end{figure*}

\barman. Figure~\ref{fig:barman_example} shows the initial and goal state for task \#1. The domain allows 12 actions: 
grasp a container, 
leave a container on the table,
fill a shot glass with an ingredient,
refill a shot glass with an ingredient,
empty a shot glass,
clean a shot glass,
pour an ingredient from a shot glass to a clean shaker,
pour an ingredient from a shot glass to a used shaker,
empty a shaker,
clean a shaker,
shake a cocktail in a shaker,
and pour from a shaker to a shot glass.

\begin{figure*}[ht]
    \centering
    \setkeys{Gin}{width=\linewidth}
    \begin{subfigure}[b]{0.1\textwidth}
    \hfil
    \includegraphics{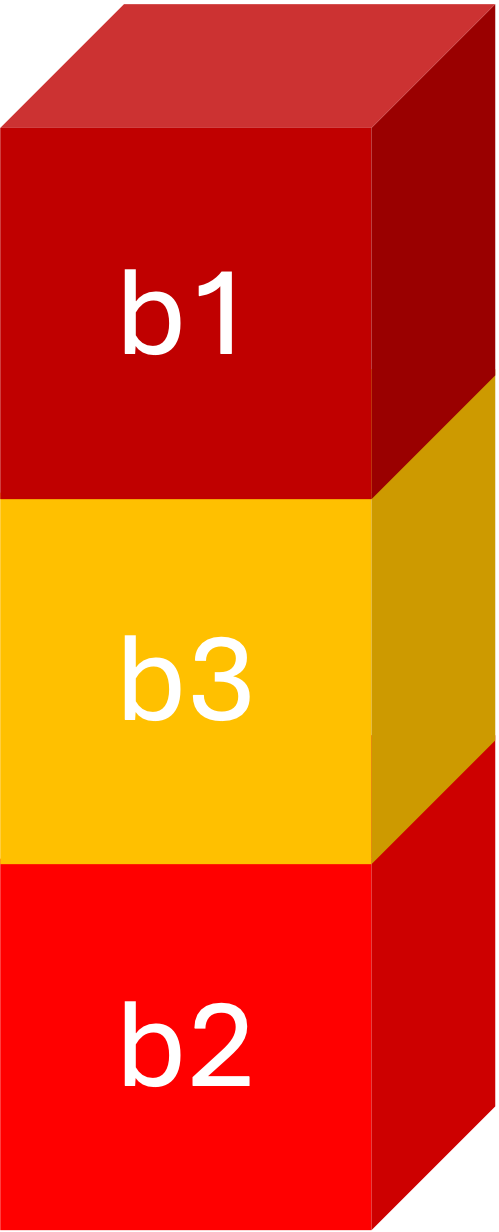}
    \caption{}
    \label{fig:bw_init}
    \end{subfigure}
    \hfil
    \begin{subfigure}[b]{0.1\textwidth}
    \includegraphics{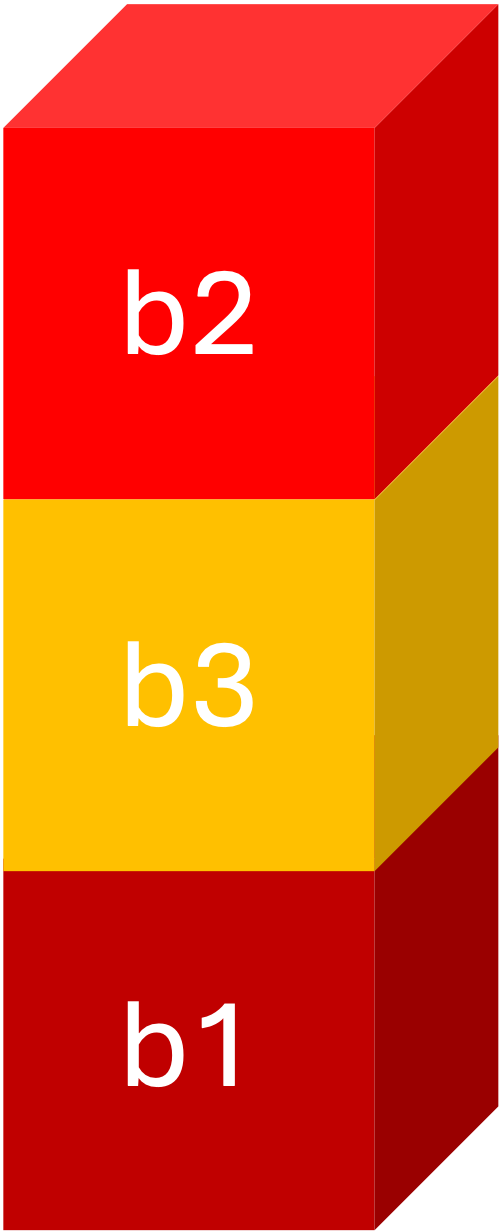}
    \caption{}
    \label{fig:bw_goal}
    \end{subfigure}
    \caption{\blocksworld task \#2 initial state (a) and goal state (b).}
    \label{fig:bw_example}
\end{figure*}

\blocksworld. Figure~\ref{fig:bw_example} shows the initial and goal state for task \#2. The domain allows 4 actions: 
pick up a block from the table,
put down a block on the table,
stack a block on top of another block,
unstack a block from on top of another block.

\begin{figure*}[ht]
    \centering
    \setkeys{Gin}{width=\linewidth}
    \begin{subfigure}[b]{0.15\textwidth}
    \hfil
    \includegraphics{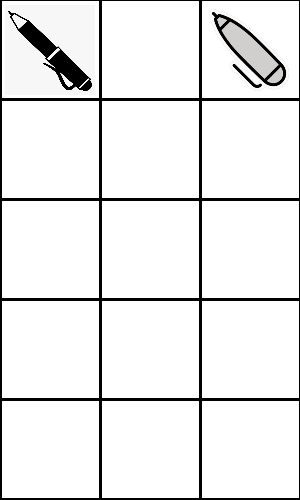}
    \caption{}
    \label{fig:floortile_init}
    \end{subfigure}
    \hfil
    \begin{subfigure}[b]{0.15\textwidth}
    \includegraphics{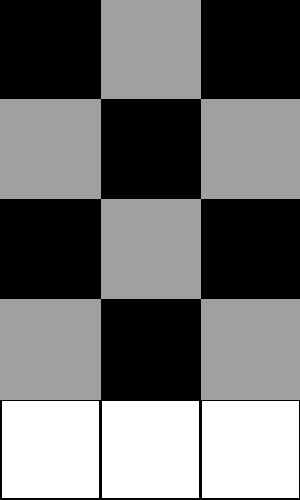}
    \caption{}
    \label{fig:floortile_goal}
    \end{subfigure}
    \caption{\floortile task \#1 initial state (a) and goal state (b).}
    \label{fig:floortile_example}
\end{figure*}

\floortile. Figure~\ref{fig:floortile_example} shows the initial and goal state for task \#1. The domain allows 7 actions: 
change the spray gun color
paint the tile that is up from the robot,
paint the tile that is down from the robot,
move up,
move down,
move right,
move left.

\begin{figure*}[ht]
    \centering
    \setkeys{Gin}{width=\linewidth}
    \begin{subfigure}[b]{0.6\textwidth}
    \hfil
    \includegraphics{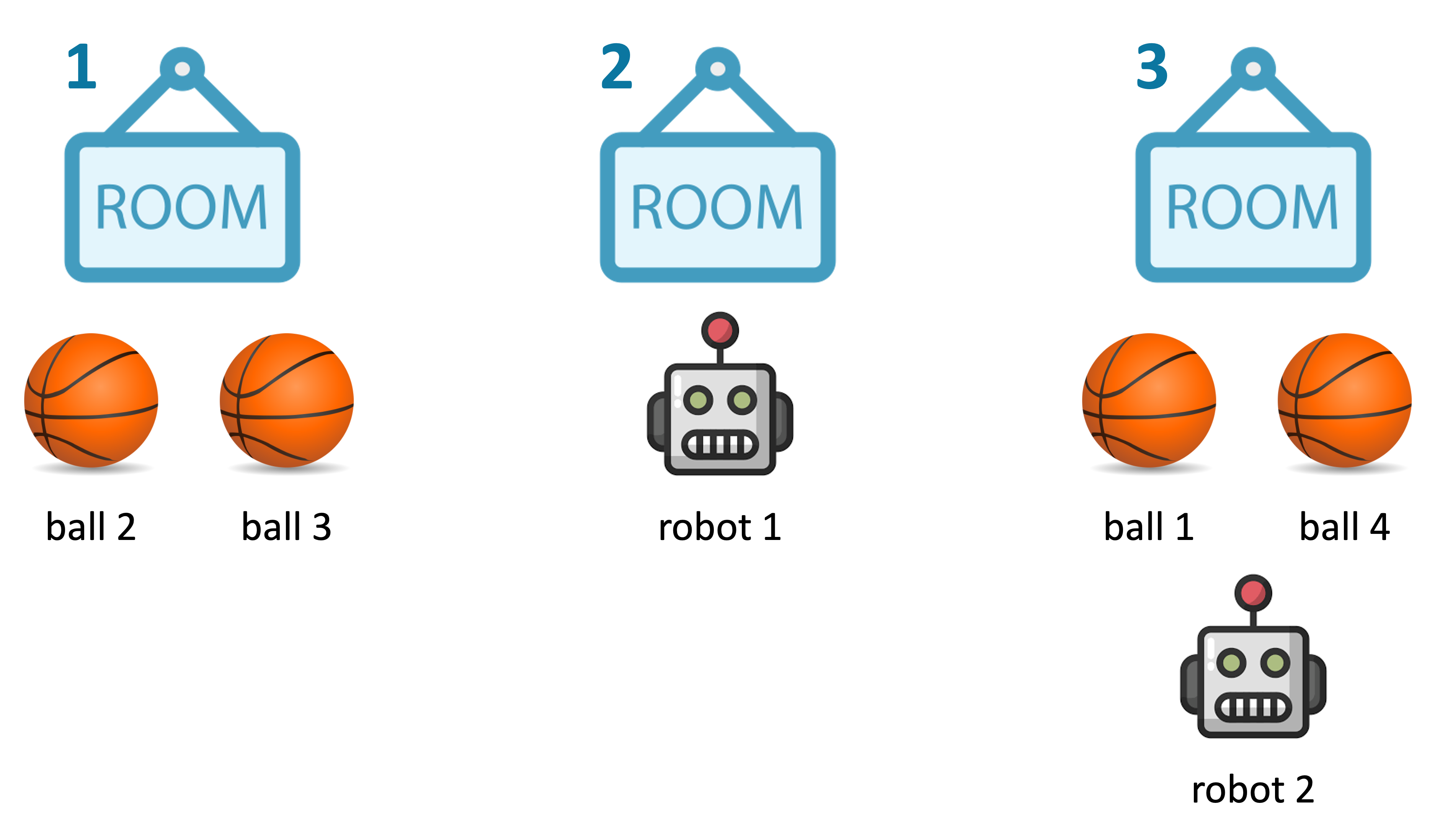}
    \caption{}
    \label{fig:grippers_init}
    \end{subfigure}
    \hspace{5mm}
    \begin{subfigure}[b]{0.6\textwidth}
    \includegraphics{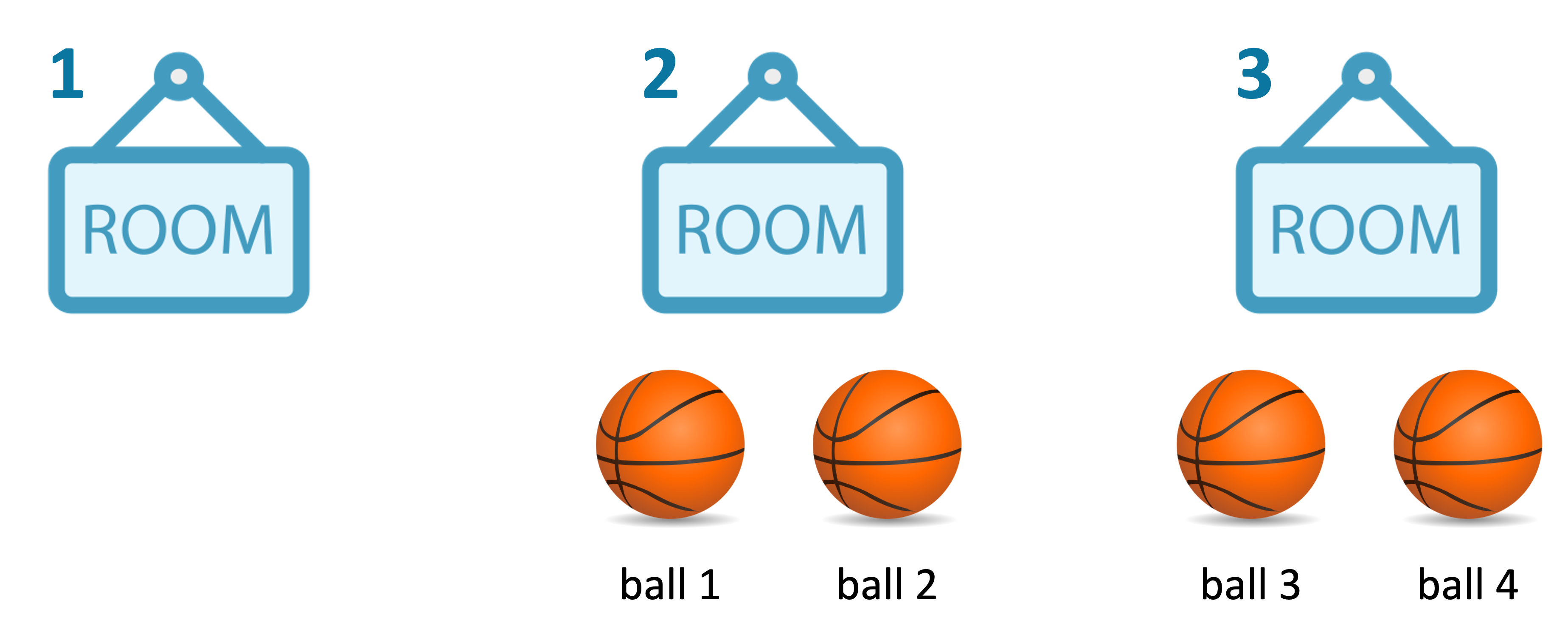}
    \caption{}
    \label{fig:grippers_goal}
    \end{subfigure}
    \caption{\grippers task \#2 initial state (a) and goal state (b).}
    \label{fig:grippers_example}
\end{figure*}

\grippers. Figure~\ref{fig:grippers_example} shows the initial and goal state for task \#2. The domain allows 3 actions: 
move from one room to another,
pick up an object using the gripper,
drop an object that it is carrying.

\begin{figure*}[ht]
    \centering
    \includegraphics[width=0.3\textwidth]{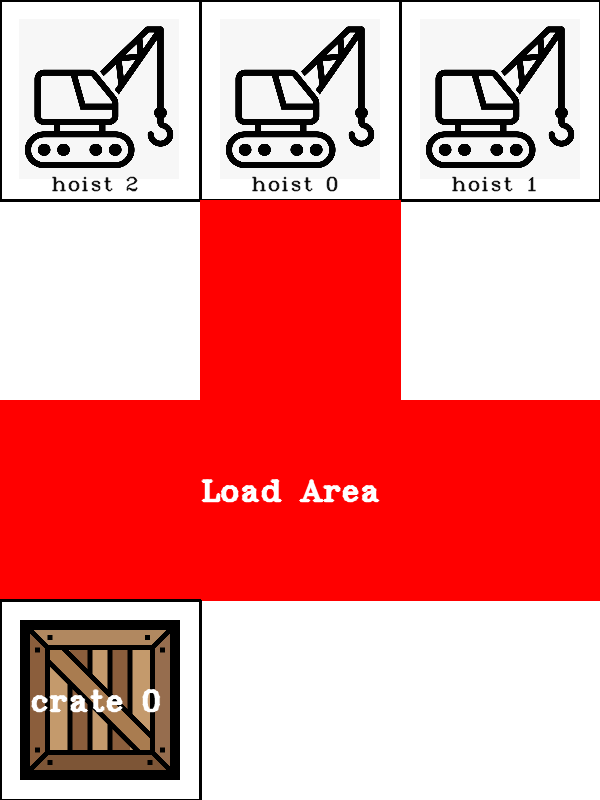}
    \caption{\storage task \#3 initial state.}
    \label{fig:storage_example}
\end{figure*}

\storage. Figure~\ref{fig:storage_example} shows the initial state for task \#3. The goal state is to move the crate to any location in the hoist store areas. The domain allows 5 actions: 
lift a crate using a hoist from a store area to an area,
drop a crate from the hoist onto a surface in a store area,
move a hoist from one store area to another connected store area,
move a hoist from a store area to a transit area,
move a hoist from a transit area to a store area.

\termes. Figure~\ref{fig:storage_example} shows the initial and goal state for task \#1. The goal state shows the height of the blocks in each position. The domain allows 7 actions: 
move from a position to another,
move up from a position to another,
move down from a position to another,
place a block at a neighboring position from the robot's current position,
remove a block at a neighboring position from the robot's current position,
create a block at the depot,
destroy a block at the depot.

\begin{figure*}[ht]
    \centering
    \includegraphics[width=\textwidth]{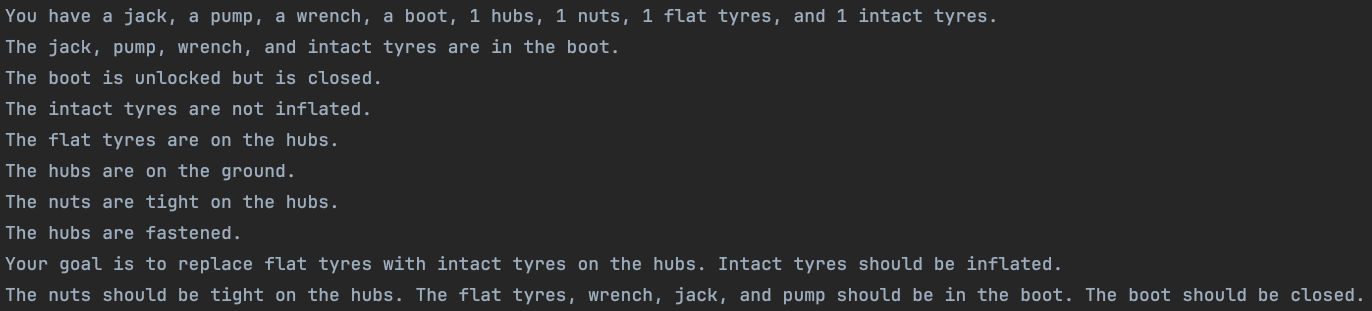}
    \caption{\tyreworld task \#1 initial state and goal state description.}
    \label{fig:tyreworld_example}
\end{figure*}

\tyreworld. Figure~\ref{fig:tyreworld_example} shows the initial and goal state description for task \#1. The domain allows 13 actions: 
open a container,
close a container,
fetch an object inside a container,
put an object into a container,
loosen a nut on a hub,
tighten a nut on a hub,
jack-up a hub,
jack-down a hub,
unfasten a nut on a hub,
fasten a nut on a hub,
remove a wheel from a hub,
put a wheel onto a hub,
inflate a wheel using a pump.

\begin{figure*}[ht]
    \centering
    \setkeys{Gin}{width=\linewidth}
    \begin{subfigure}[b]{0.15\textwidth}
    \hfil
    \includegraphics{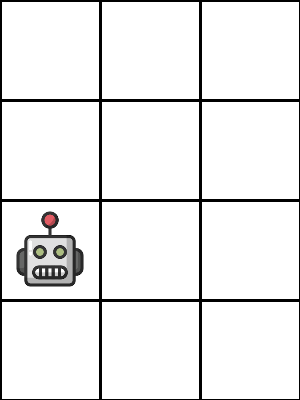}
    \caption{}
    \label{fig:termes_init}
    \end{subfigure}
    \hfil
    \begin{subfigure}[b]{0.15\textwidth}
    \includegraphics{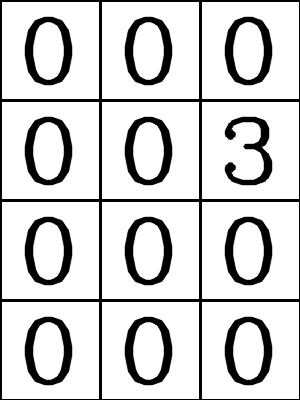}
    \caption{}
    \label{fig:termes_goal}
    \end{subfigure}
    \caption{\termes task \#1 initial state (a) and goal state (b).}
    \label{fig:termes_example}
\end{figure*}

\subsection{Other Hyperparameters and Details}
The total cost of the API calls of \ourmethodshort over 7 domains is around \$80 for 1 run.
We do not put the pure random baselines in Figure~\ref{fig:main_results}, because none of the pure random baselines complete any task within 1k iterations.

Besides, we only use the cheapest configuration of \planner, \ie, we are not using very smart symbolic solver. \planner has various advanced search configurations including using A* search (\texttt{seq-opt-lmcut}), using heuristics plus hill climbing algorithms (\texttt{seq-sat-fd-autotune-1}).
We use the \texttt{lama-first} configuration, which is close to lazy greedy search designed to find solutions quickly without much regard for plan cost.

The IPC domains and tasks~\cite{seipp-et-al-zenodo2022} are licensed under the MIT License. The \planner planner is distributed under the GNU GPLv2 license, while the \validator validator is licensed under the BSD 3-Clause, with copyright held by the University of Strathclyde, King's College London, and Schlumberger Ltd. Our use of the dataset and software complies with the terms of their respective licenses.

\section{Ablation studies}
We perform ablation studies on prospection steps, number of sampled trajectories, number of candidate trajectories, and number of failed trajectories per run, to show their influence on the \ourmethodshort framework.
The results validate our choice of the hyperparameters in the main results.

\begin{figure*}[ht]
    \centering
    \setkeys{Gin}{width=\linewidth}
    \begin{subfigure}[b]{0.245\textwidth}\includegraphics{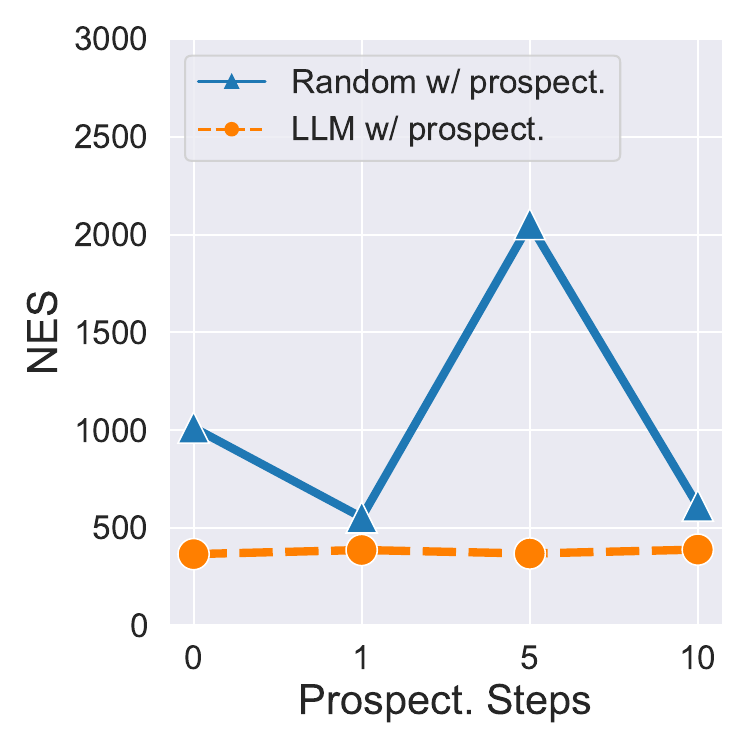}
    \caption{}
    \label{fig:prospect}
    \end{subfigure}
    \hfil
    \begin{subfigure}[b]{0.245\textwidth}
    \includegraphics{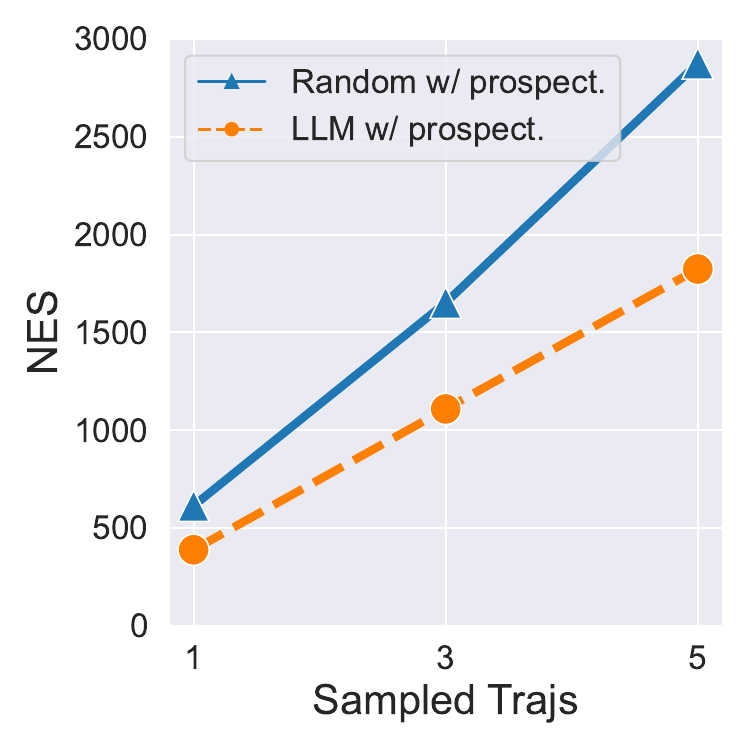}
    \caption{}
    \label{fig:sampled-trajs}
    \end{subfigure}
    \begin{subfigure}[b]{0.245\textwidth}
    \hfil
    \includegraphics{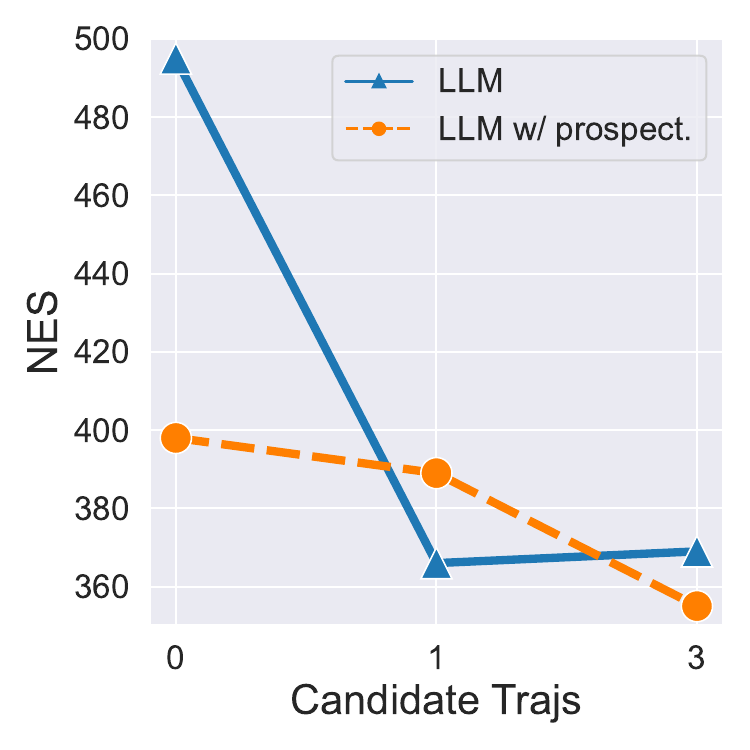}
    \caption{}
    \label{fig:candid-trajs}
    \end{subfigure}
    \hfil
    \begin{subfigure}[b]{0.245\textwidth}
    \includegraphics{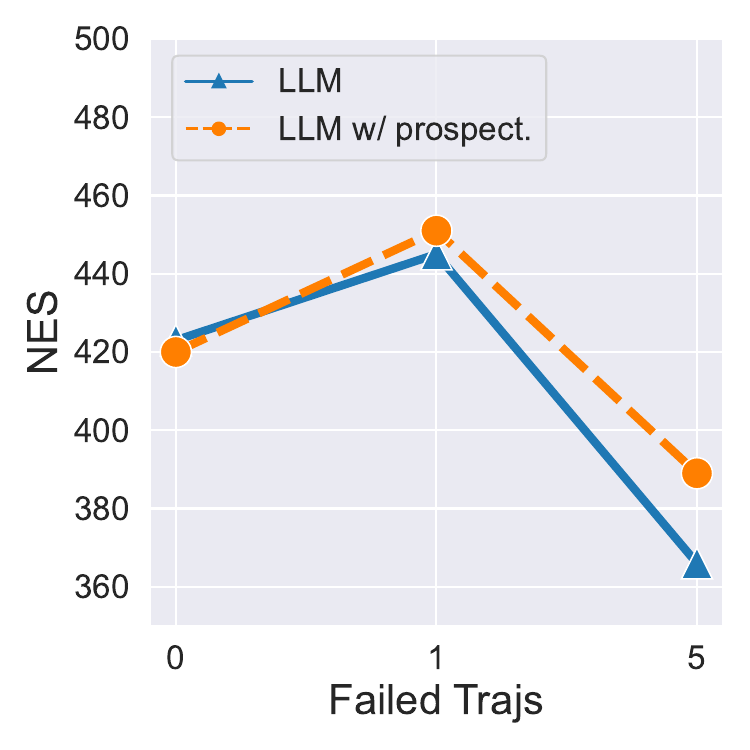}
    \caption{}
    \label{fig:failed-trajs}
    \end{subfigure}
    \caption{Ablation studies for \ourmethodshort. (a). Varying the number of prospection steps $v$. Few prospection steps is enough for \ourmethodshort. (b). Varying the number of sampled trajectories $l$. One sampled trajectory per run is enough. 
    (c). Varying the number of candidate trajectories $k$ passed from the symbolic solver to the LLM trajectory sampler. More candidate trajectories help when prospection is used. 
    (d). Varying the number of failed trajectories $g$ shown to the LLM trajectory sampler. 
    A certain amount of failed trajectories is required for the LLM.}
    \label{fig:ablations}
\end{figure*}

\paragraph{Few prospection steps is enough for \ourmethodshort.}
We vary prospection steps $v$ from $0,1,5,10$ in Figure~\ref{fig:prospect}.
Notice the random baseline does not find a solution when $v=0,1,5$, and reaches an Acc of 78.5.
We conclude the number of prospection steps matters more when the trajectory sampler is random, while a small number of prospection is enough for language model trajectory sampler.

\paragraph{One sampled trajectory per run is enough.}
We vary the number of sampled trajectories $l$ from $1,3,5$ in Figure~\ref{fig:sampled-trajs}.
For both LLM and random samplers, We see the total number of steps grows linearly with number of sampled trajectory, which means LLMs are hard to learn more information from just sampling more each run, and one sampled trajectory per run is enough.

\paragraph{More candidate trajectories help on prospection.}
We vary the number of candidate trajectories $k$ from $0,1,3$ in the LLM prompt. 
The results in Figure~\ref{fig:candid-trajs} suggest including candidate trajectory in the prompt is beneficial, while  one candidate trajectory is enough for pure LLM prediction, more candidate trajectories help LLM with prospection.
On the other hand, as the candidate trajectories can be very long for certain domain like \barman, and thus result in a very long prompt, we do not experiment with more than 3 candidate trajectories in the prompt. 

\paragraph{Choose the number of failed trajectories with care.}
We vary the number of failed trajectories $g$ from $0,1,5$ in the LLM prompt. 
To select $g$ failed trajectories, we first filter trajectories with no less than 3 steps, and then sample $g$ trajectories from the filtered ones for the LLM prompt to avoid generating those trajectories as prefix.
The results in Figure~\ref{fig:failed-trajs} suggest that multiple failed trajectories help LLM action semantics prediction, but only one failed trajectory could be harmful.
We hypothesize the one failed trajectory in the input might distract the model from following the candidate trajectories.

\section{LLM Prompts}
\label{sec:app_prompt}

\begin{figure*}[ht]
    \centering
    \includegraphics[width=\textwidth]{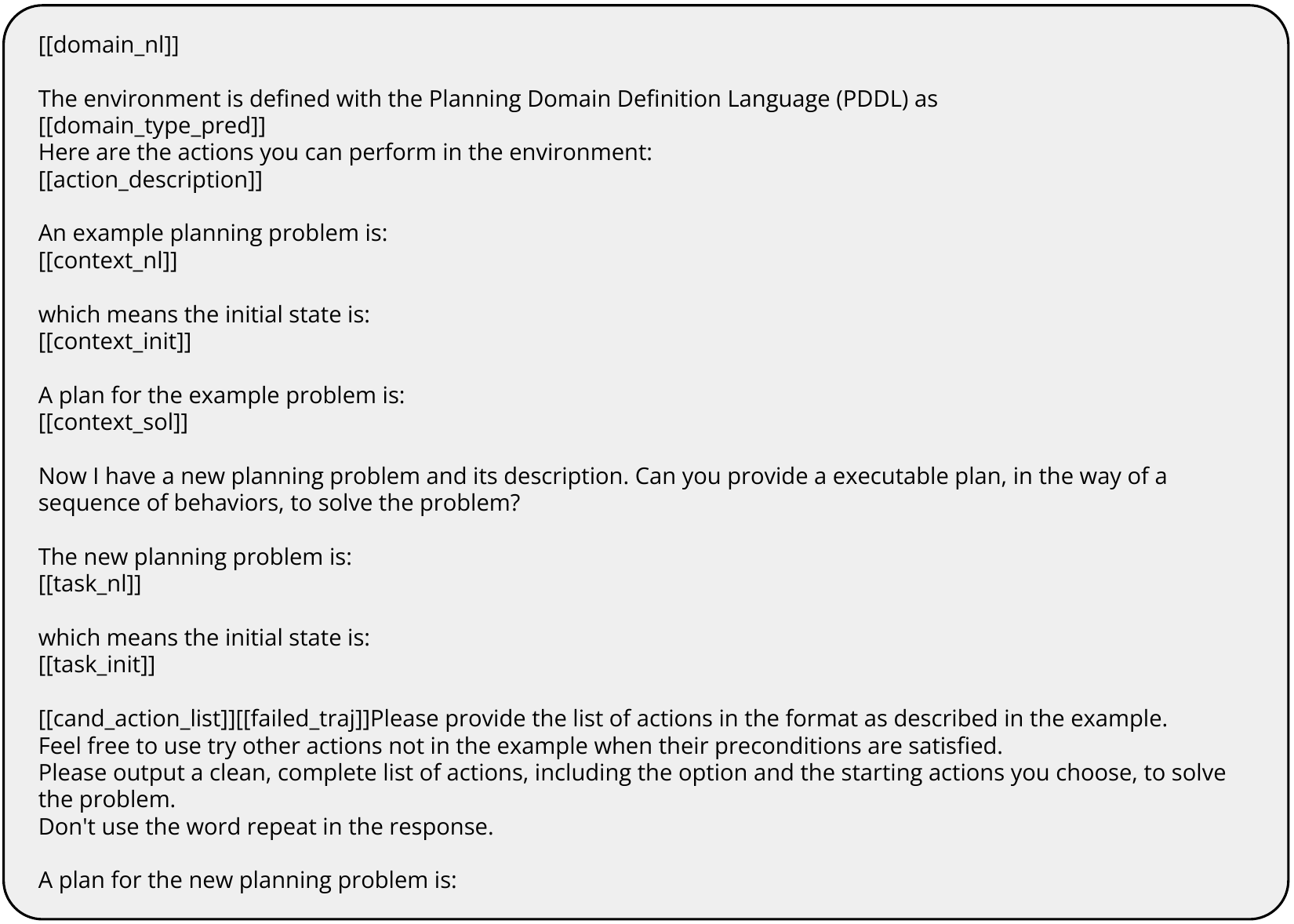}
    \caption{LLM trajectory sampler prompt template}
    \label{fig:traj_sampler_prompt_template}
\end{figure*}

\begin{figure*}[ht]
    \centering
    \includegraphics[width=\textwidth]{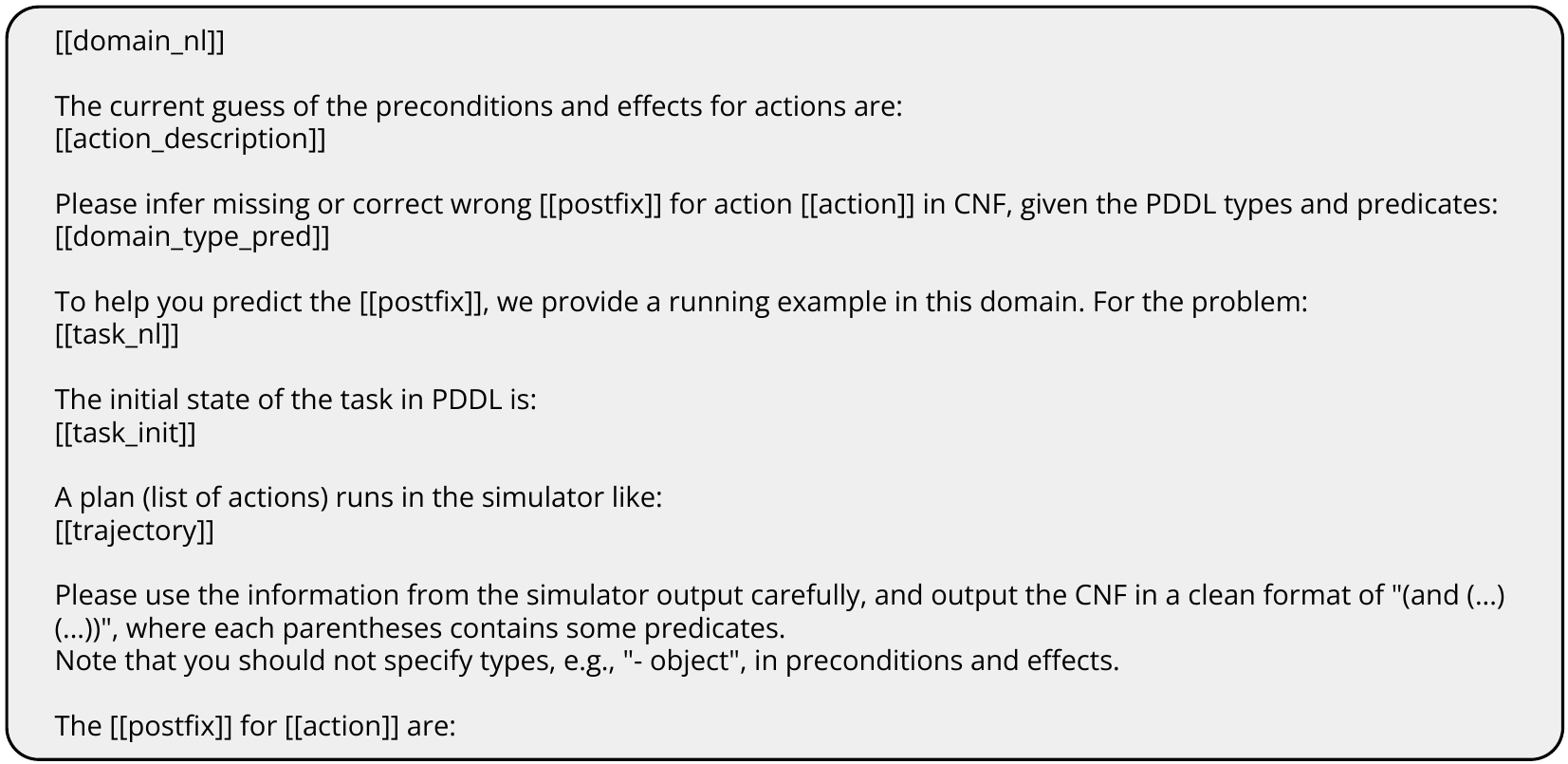}
    \caption{LLM action semantics predictor prompt template}
    \label{fig:asp_prompt_template}
\end{figure*}

\begin{figure*}[t]
    \centering
    \includegraphics[width=\textwidth]{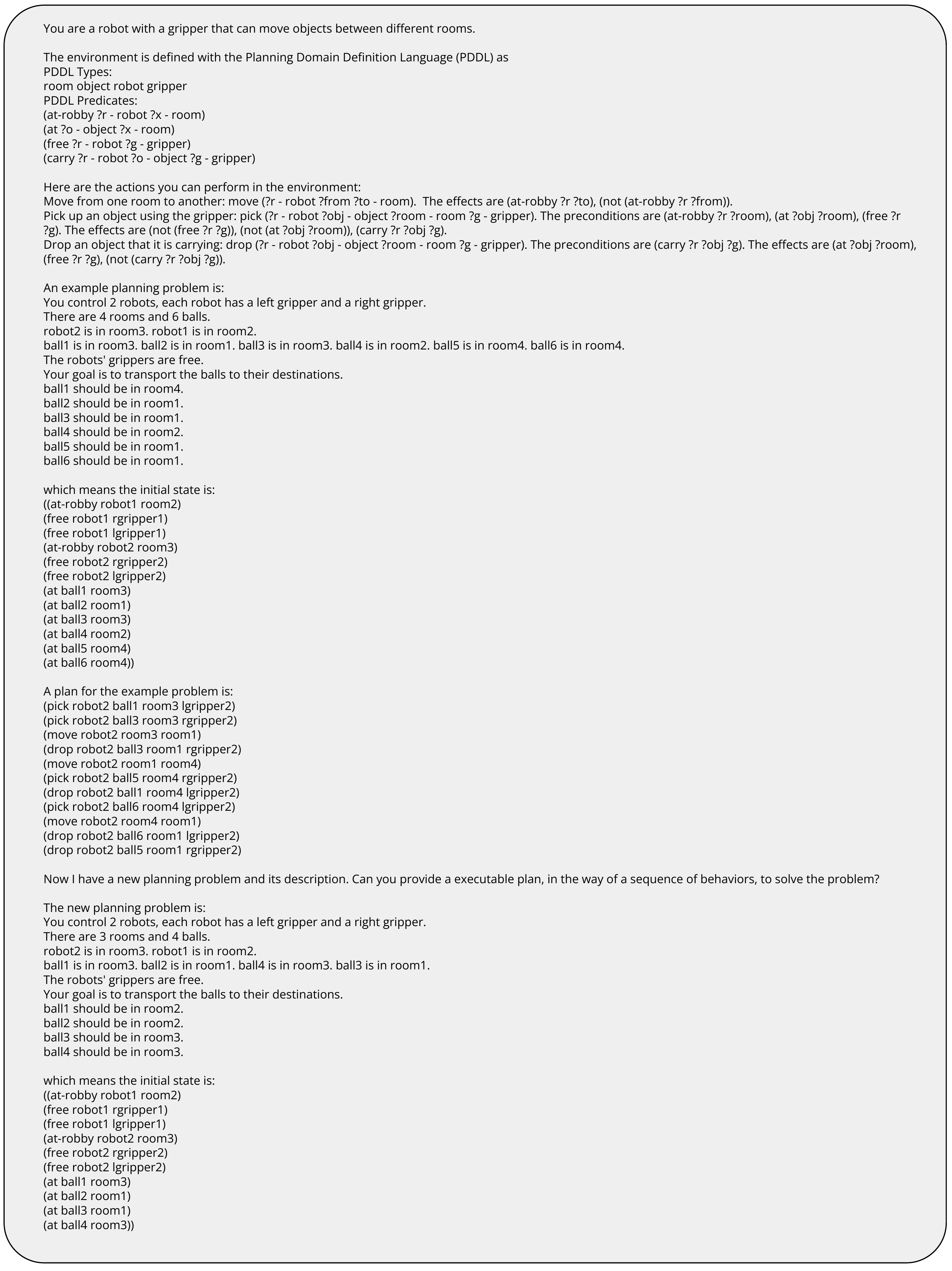}
    \caption{LLM trajectory sampler prompt example}
    \label{fig:traj_sampler_prompt_template_1}
\end{figure*}

\begin{figure*}[t]
    \centering
    \includegraphics[width=\textwidth]{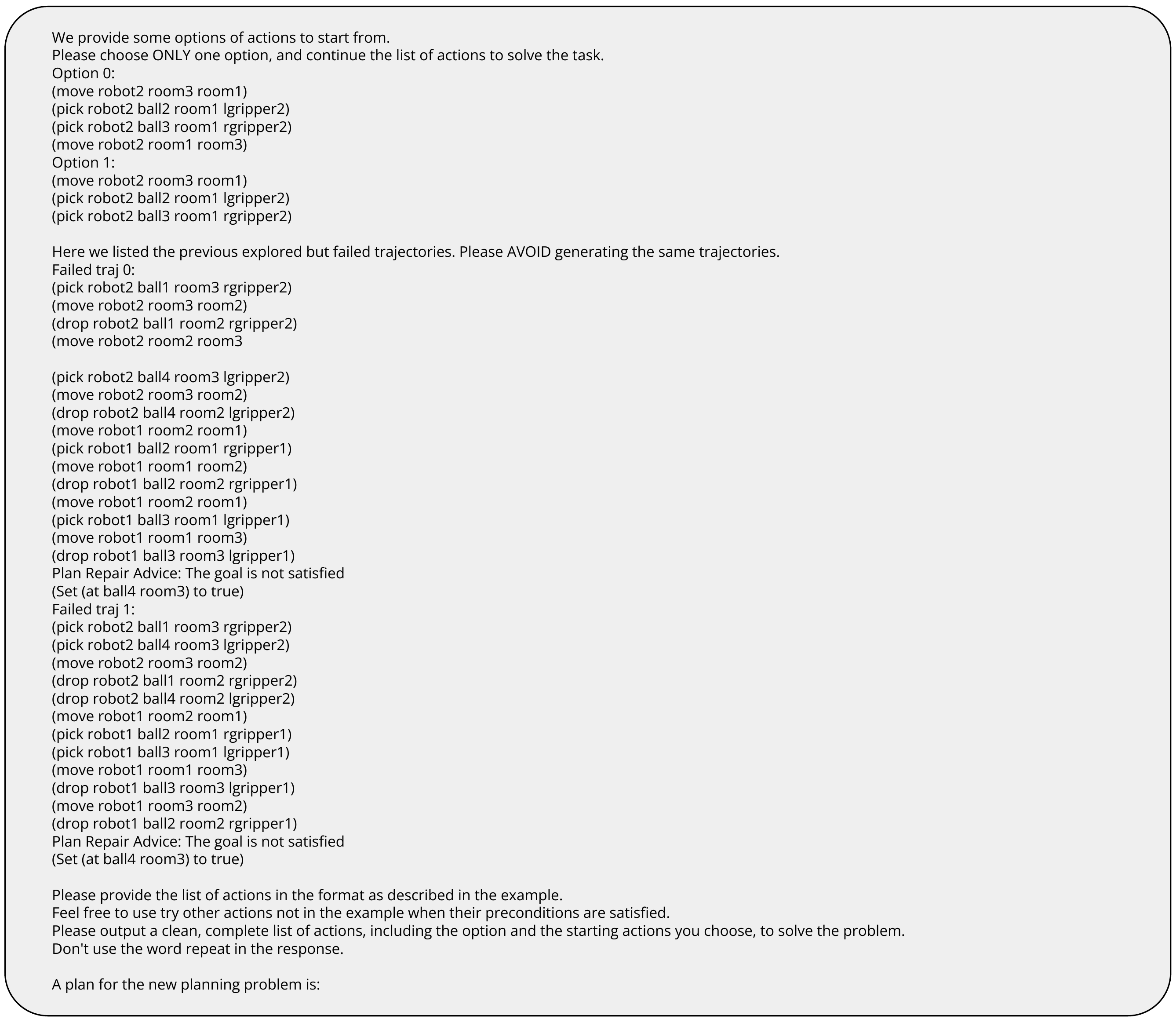}
    \caption{LLM trajectory sampler prompt example (cont')}
    \label{fig:traj_sampler_prompt_template_2}
\end{figure*}

\begin{figure*}[t]
    \centering
    \includegraphics[width=\textwidth]{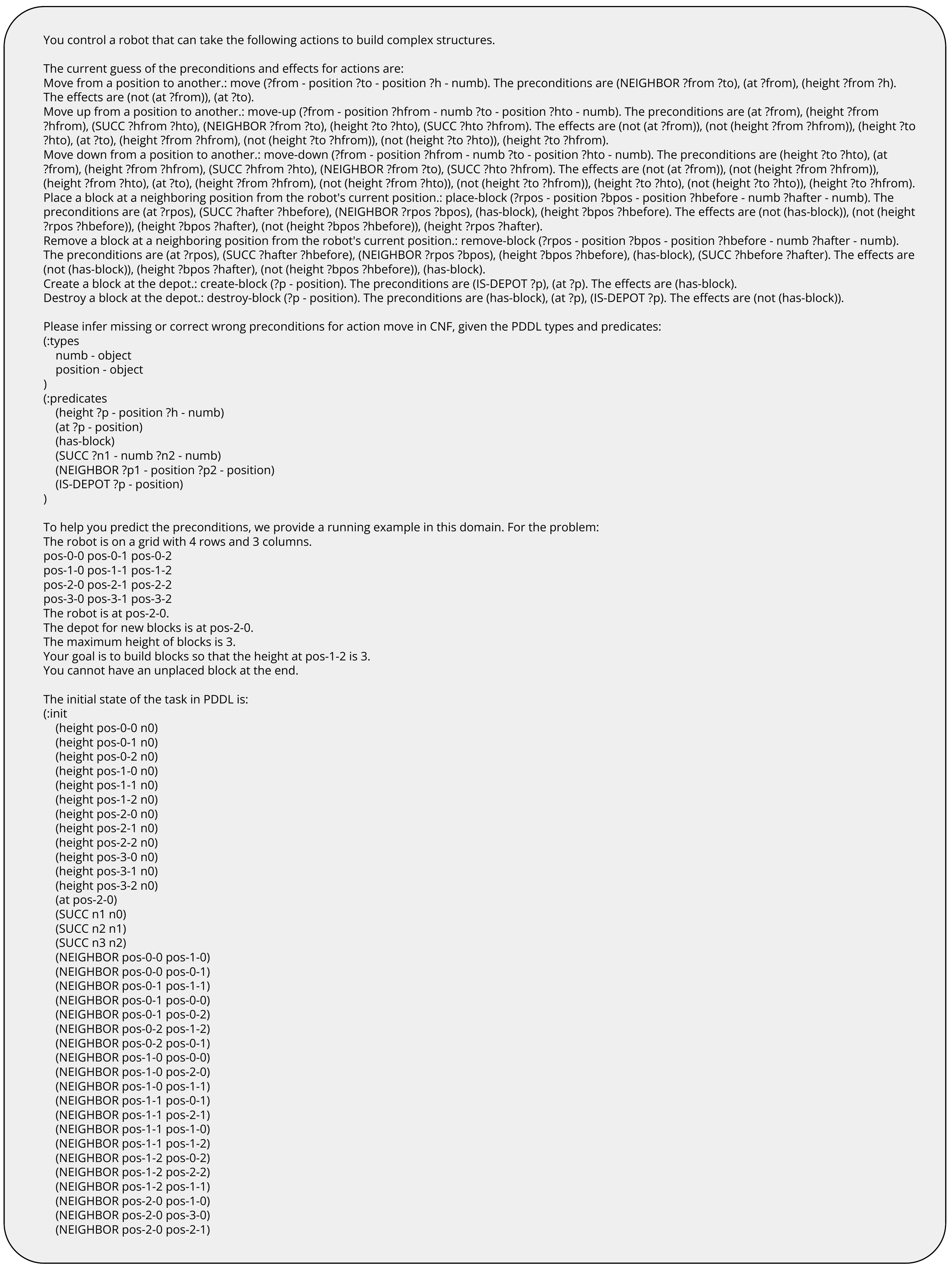}
    \caption{LLM action semantics predictor prompt example}
    \label{fig:asp_prompt_example_1}
\end{figure*}

\begin{figure*}[t]
    \centering
    \includegraphics[width=\textwidth]{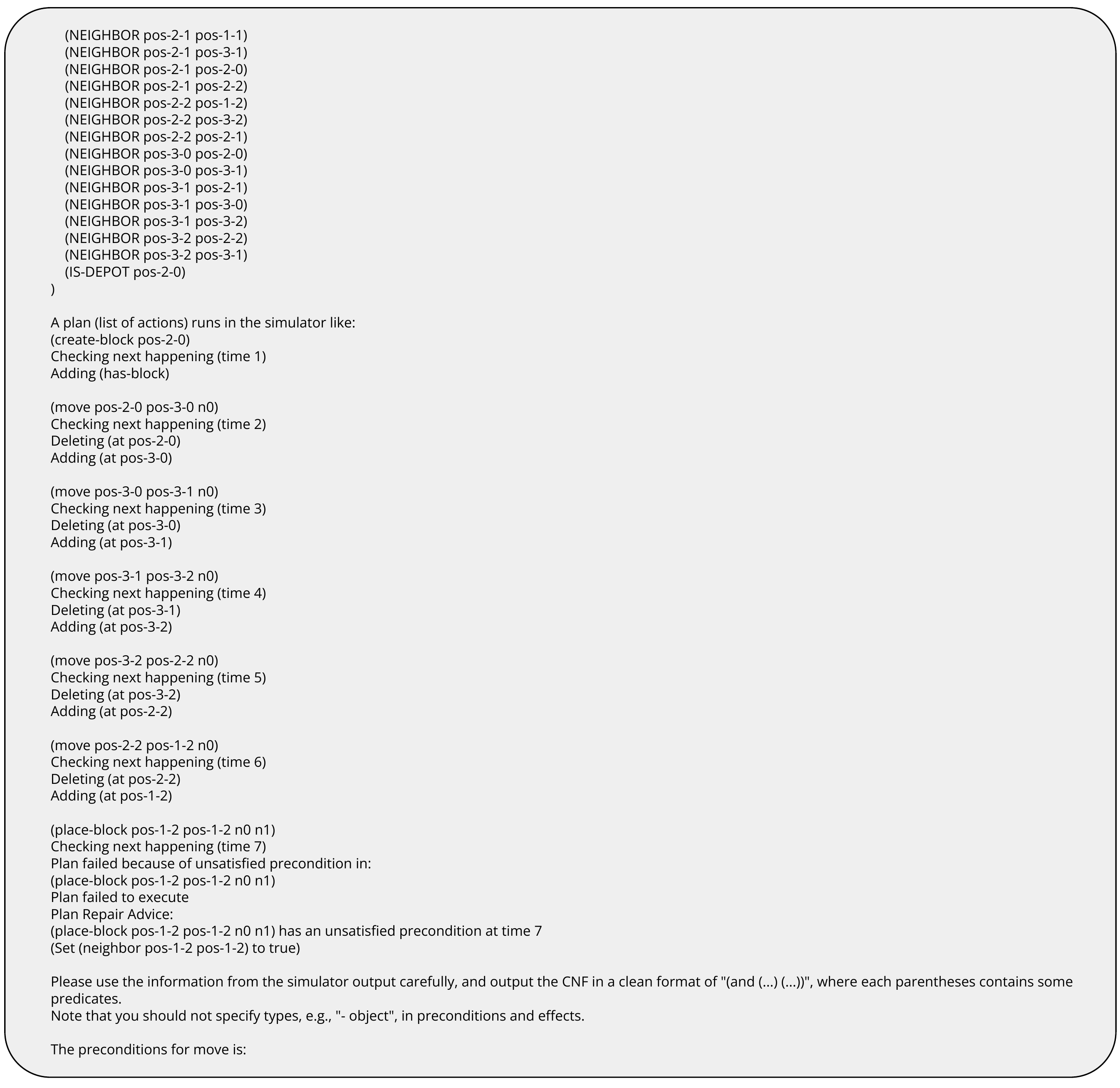}
    \caption{LLM action semantics predictor prompt example (cont')}
    \label{fig:asp_prompt_example_2}
\end{figure*}

We list the prompt template for LLM trajectory sampler (Figure~\ref{fig:traj_sampler_prompt_template}) and LLM action semantics predictor (Figure~\ref{fig:asp_prompt_template}), and provide one complete example for each.
Figure~\ref{fig:traj_sampler_prompt_template_1},~\ref{fig:traj_sampler_prompt_template_2} is the example for trajectory sampler, and Figure~\ref{fig:asp_prompt_example_1},~\ref{fig:asp_prompt_example_2} is the example for action semantics predictor.

\end{document}